\documentclass[times,twocolumn,final]{elsarticle}
\usepackage{graphicx,stfloats,times,amssymb,amsmath,subfigure,url,multirow,bm,color,booktabs,hyperref}
\usepackage[flushleft]{threeparttable}



\graphicspath{{fig/}}
\journal{ }
\begin{document}

\begin{frontmatter}

\title{Gated Parametric Neuron for Spike-based Audio Recognition}
\author[]{Haoran Wang}\ead{hrwang21@hust.edu.cn}
\author[]{Herui Zhang}\ead{heruizhang@hust.edu.cn}
\author[]{Siyang Li}\ead{syoungli@hust.edu.cn}
\author[]{Dongrui Wu\corref{cor1}}\ead{drwu@hust.edu.cn}

\cortext[cor1]{Corresponding author}

\address{Key Laboratory of the Ministry of Education for Image Processing and Intelligent Control, School of Artificial Intelligence and Automation, Huazhong University of Science and Technology, Wuhan 430074, China}

\begin{abstract}
Spiking neural networks (SNNs) aim to simulate real neural networks in the human brain with biologically plausible neurons. The leaky integrate-and-fire (LIF) neuron is one of the most widely studied SNN architectures. However, it has the vanishing gradient problem when trained with backpropagation. Additionally, its neuronal parameters are often manually specified and fixed, in contrast to the heterogeneity of real neurons in the human brain. This paper proposes a gated parametric neuron (GPN) to process spatio-temporal information effectively with the gating mechanism. Compared with the LIF neuron, the GPN has two distinguishing advantages: 1) it copes well with the vanishing gradients by improving the flow of gradient propagation; and, 2) it learns spatio-temporal heterogeneous neuronal parameters automatically. Additionally, we use the same gate structure to eliminate initial neuronal parameter selection and design a hybrid recurrent neural network-SNN structure. Experiments on two spike-based audio datasets demonstrated that the GPN network outperformed several state-of-the-art SNNs, could mitigate vanishing gradients, and had spatio-temporal heterogeneous parameters. Our work shows the ability of SNNs to handle long-term dependencies and achieve high performance simultaneously.
\end{abstract}


\end{frontmatter}

\section{Introduction} \label{sect:Intro}

Spiking neural networks (SNNs), known as the `third generation neural networks' \cite{maass1997networks}, are brain-inspired models to simulate the dynamics of the human brain \cite{Gerstner2014Neuronal}. Unlike artificial neural networks (ANNs) that are widely used in deep learning, SNNs employ spiking neurons to output binary spikes rather than continuous values. More specifically, SNNs utilize the firing rate and the firing time information to process spatio-temporal information \cite{wu2018spatio}. Due to their sparse asynchronous working mechanism, SNNs can better handle event-based data \cite{deng2020rethinking,li2017cifar10,orchard2015converting} and reduce the network energy consumption \cite{cao2015spiking,kim2020spiking,Shen2023ESL}. They have been used for image and audio classification \cite{shrestha2018slayer,Xu2022Hierarchical,XU2020Deep,Xu2023Constructing}, brain mapping and understanding \cite{kasabov2014neucube}, robot control \cite{bing2018survey,lobov2020spatial}, etc. SNNs also facilitate the development of neuromorphic computing hardware \cite{davies2018loihi,debole2019truenorth}.

Backpropagation with surrogate gradients is widely used to train SNNs. Differentiable surrogate functions are used to replace the non-differentiable Heaviside function in determining spike emission during backpropagation \cite{netfci2019surrogate}. The surrogate functions can assume various shapes, e.g., triangular \cite{bellec2018long,Rathi2021diet}, fast sigmoid \cite{cramer2022heidelberg}, arc tangent \cite{fang2021deep,fang2021incorporating}, etc. With surrogate functions, SNNs can be trained directly, similar to ANNs.

The leaky integrate-and-fire (LIF) neuron is a simple phenomenological spiking neuron that focuses on the specific spiking time. Due to its simplicity, LIF is frequently used for neural coding and network dynamics modeling \cite{Gerstner2014Neuronal}. However, LIF has the following limitations:

\begin{enumerate}
\item \emph{Vanishing gradients.} Existing SNNs, similar to the classical recurrent neural networks (RNNs) \cite{bengio1994learning}, have difficulty accommodating long-term dependencies in the time domain. Ponghiran and Roy \cite{ponghiran2022spiking} analyzed the vanishing gradient problem of the current-based LIF (Cuba-LIF) network \cite{dampfhoffer2022investigating,perez2021neural}, which has two recurrent states of synaptic current and membrane potential. They showed that the gradients mainly flow through the synaptic current pathway, and introduced a gating mechanism and multi-bit outputs to modify the inherent recurrent dynamics of the Cuba-LIF network. This paper also analyzes the vanishing gradient problem, but we focus on the single membrane potential pathway for LIF without a synaptic current pathway.

\item \emph{Fixed neuronal parameters.} The neuronal parameters (time constant and firing threshold) of LIF neurons are usually fixed \cite{yao2021temporal,deng2022temporal,yu2022STSC}. In contrast, these electrophysiological properties of real neurons in the human and animal nervous systems are observed to be heterogeneous \cite{perez2021neural,connors1982electrophysiological,degenetais2002electrophysiological,pennartz1998electrophysiological}. A group of middle temporal gyrus spiny neurons in the human brain from the Allen Brain Atlas dataset \cite{allen2015} were chosen for illustration. Figure~\ref{fig_begin} shows the firing thresholds under three different input current stimuli (ramp, long square, and short square). The firing thresholds of different neurons are different. For temporal dynamics, experiments have shown that the firing thresholds are also dynamically variable \cite{azouz2000dynamic}. In summary, the firing thresholds vary among different neurons (spatial domain) and at different times (temporal domain).

    \begin{figure*}[htpb]    \centering
    \includegraphics[width=\linewidth,clip]{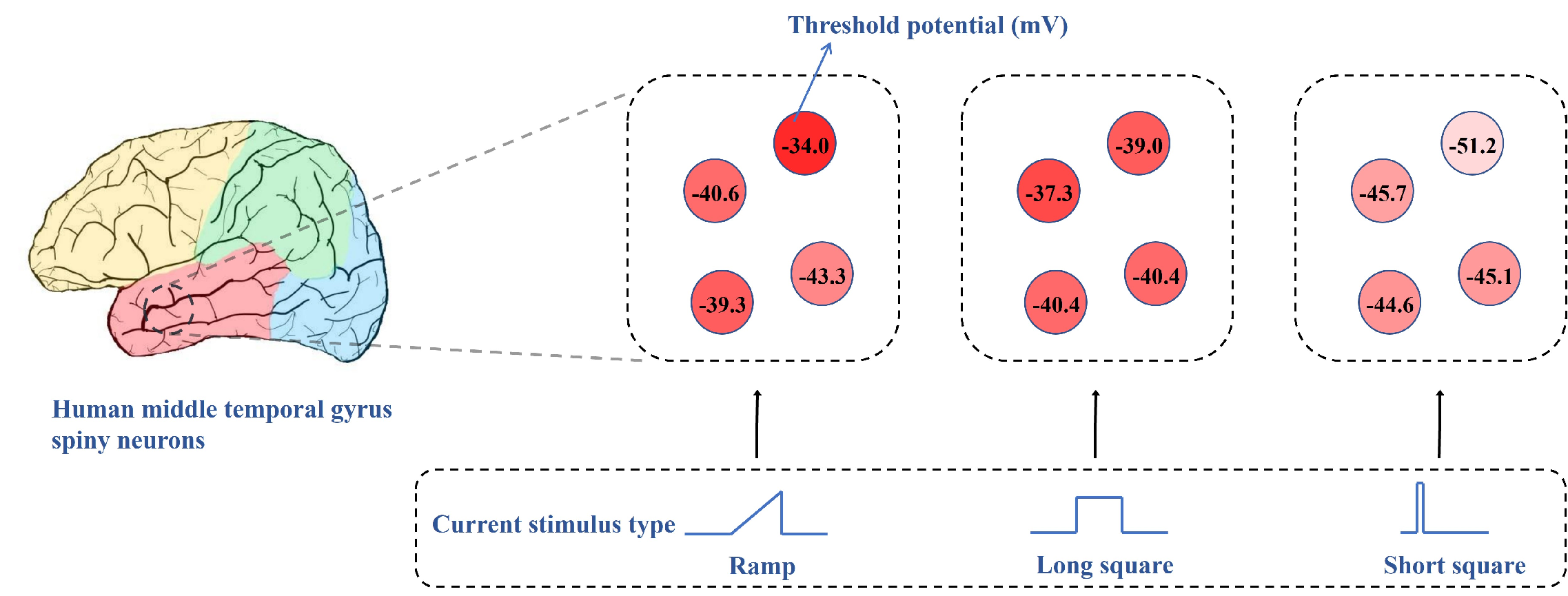}
    \caption{Firing thresholds of the human brain's middle temporal gyrus spiny neurons. The firing thresholds of different neurons are different for the same input current type. The color depth indicates the threshold value.} \label{fig_begin}
    \end{figure*}

\item \emph{Manual specification of neuronal parameters.} For LIF models with fixed \cite{wu2018spatio} or variable neuronal parameters \cite{fang2021incorporating,yin2021accurate}, their initial values are usually manually specified. However, it is challenging to find the optimal values, and time-consuming trial-and-errors may be needed \cite{polap2022heuristic}. A model without manual specification of neuronal parameters will be convenient.

\item \emph{Performance degradation.} Due to the complex neuronal dynamics and binary outputs, the performance of SNNs is usually worse than ANNs in ANN-oriented tasks \cite{deng2020rethinking}. Some studies tried to improve the performance of SNNs in backpropagation training, by introducing classical ANN techniques to them, e.g., residual learning \cite{fang2021deep,lee2020enabling}, attention mechanism \cite{yao2021temporal,yu2022STSC}, non-binary outputs \cite{wu2021liaf}, hybrid ANN-SNN structure \cite{Kosta_2023_live,Zhang_2021_event}, etc.
\end{enumerate}

This paper proposes a gated parametric neuron (GPN) to cope with the above limitations simultaneously. GPN adds the following features to LIF with simple gating structures:
\begin{enumerate}
    \item \emph{Mitigation of the vanishing gradients.} We first analyze the pattern of vanishing gradients in LIF networks. The gradients in the temporal domain vanish along the membrane potential pathway. Therefore, we consider the gating structure used by Long Short-Term Memory (LSTM) \cite{Hochreiter1997LSTM} and Gated Recurrent Unit (GRU) \cite{cho2014GRU}, which can effectively mitigate the vanishing gradient problem of vanilla RNNs. Compared with previous works \cite{ponghiran2022spiking,dampfhoffer2022investigating}, we use a simpler LIF model with membrane potential and synaptic current as the gate inputs, and the gate outputs as the membrane leaky factors. A forget gate and an input gate, similar to LSTM, are added to mitigate the vanishing gradients of the LIF network.
    \item \emph{Spatio-temporal neuronal parameters.} For the gating mechanism, the gate outputs are dynamically variable across different neurons and at different times. In addition to the gates to produce the membrane leaky factors, we also add a threshold gate to compute the firing thresholds. Therefore, neuronal parameters (membrane time constant and firing threshold) achieve heterogeneities at different neurons (spatial domain) and at different times (temporal domain).
    \item \emph{No manual specifications of neuronal parameters.} The neuronal parameters are calculated with the membrane potential and synaptic current at each time step through the gates, eliminating the manual specification of neuronal parameters.
    \item \emph{Hybrid RNN-SNN structure.}  GPN achieves the advantages of RNNs with gating mechanisms. A bypass gate enables the transmission of information on membrane potentials of adjacent moments. Therefore, the information on the membrane potential pathway will flow simultaneously along the original and the bypass ways, accomplishing a hybrid RNN-SNN structure.
\end{enumerate}

The remainder of this paper is organized as follows: Section~\ref{sect:Related} describes related works. Section~\ref{sect:Methods} introduces LIF and our proposed GPN model. Section~\ref{sect:Experiments} describes the datasets and detailed experimental settings. Section~\ref{sect:Results} presents experimental results. Finally, Section~\ref{sect:Conclusions} draws conclusions. The source code is available on GitHub\footnote{https://github.com/hrje123/hr-GPN-SNN}.

\section{Related Works} \label{sect:Related}

\subsection{Trainable parameters}

Multiple approaches \cite{Rathi2021diet,fang2021incorporating,perez2021neural} have used backpropagation to train the neuronal parameters directly. Fang \emph{et al.} \cite{fang2021incorporating} proposed a LIF-based model to learn membrane time constants, which vary among different layers of the SNN. Perez-Nieves \emph{et al.} \cite{perez2021neural} learned the membrane and synaptic time constant for each neuron, improving the spatial heterogeneity of the network. Rathi and Roy \cite{Rathi2021diet} trained both the membrane leaky factors and the firing thresholds. However, the trainable parameters are updated only during backpropagation, lacking temporal heterogeneity during forward propagations in an iteration. Additionally, their performance is affected by the initial manual specification \cite{fang2021incorporating} and suffers from the vanishing gradient problem \cite{ponghiran2022spiking}.

\subsection{Adaptive thresholds}

Neurons with adaptive thresholds have also been studied extensively \cite{bellec2018long,yin2021accurate}. When a neuron fires, the firing threshold potential rises. Otherwise, it decays back to its stationary value. Bellec \emph{et al.} \cite{bellec2018long} added the adaptive threshold mechanism to enrich the inherent neuronal dynamics. Yin \emph{et al.} \cite{yin2021accurate} combined adaptive thresholds with trainable membrane time constants. Our analysis in Section~\ref{sect:vanish} shows that neurons with fixed membrane time constants cannot mitigate the vanishing gradient problem of the LIF network. The adaptive dynamic enables heterogeneous thresholds but demands additional hyper-parameters.

\subsection{Gating mechanism}

The gating mechanism has been used to improve neuronal dynamics and mitigate the vanishing gradients for the Cuba-LIF network \cite{ponghiran2022spiking,dampfhoffer2022investigating,Xu2023enhancing}. Ponghiran and Roy \cite{ponghiran2022spiking} used gates to update the synaptic currents while fixing the equations for membrane potential updating. Dampfhoffer \emph{et al.} \cite{dampfhoffer2022investigating} integrated Cuba-LIF and GRU structures, achieving better performance with the reduced number of operations. Yao \cite{yao2022glif} proposed a heterogeneous and adaptable gating mechanism to fuse multiple biological features. Inspired by the gating mechanism, we propose a hybrid RNN-SNN model to mitigate vanishing gradients. Moreover, it generates spatio-temporal heterogeneous neuronal parameters without manual specification.

\section{Method} \label{sect:Methods}

\subsection{Leaky integrate-and-fire (LIF) neuron}

\begin{figure}[htpb]    \centering
    \subfigure[]{\includegraphics[width=.90\linewidth,clip]{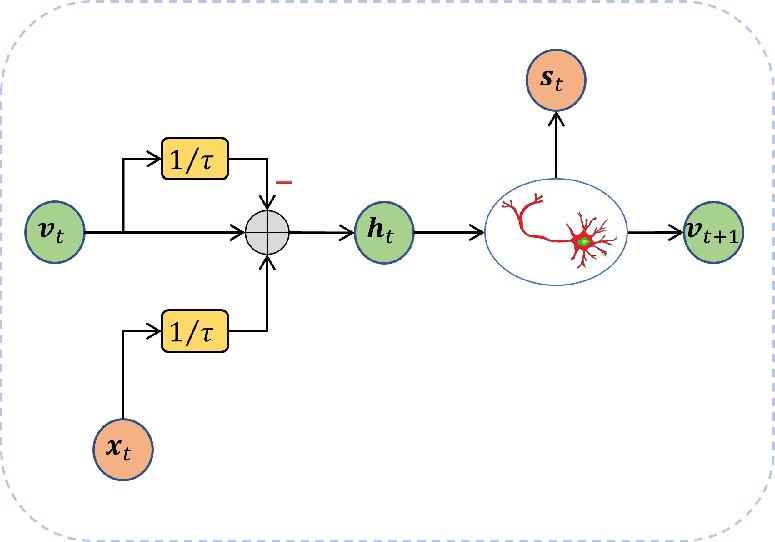}}
    \subfigure[]{\includegraphics[width=.95\linewidth,clip]{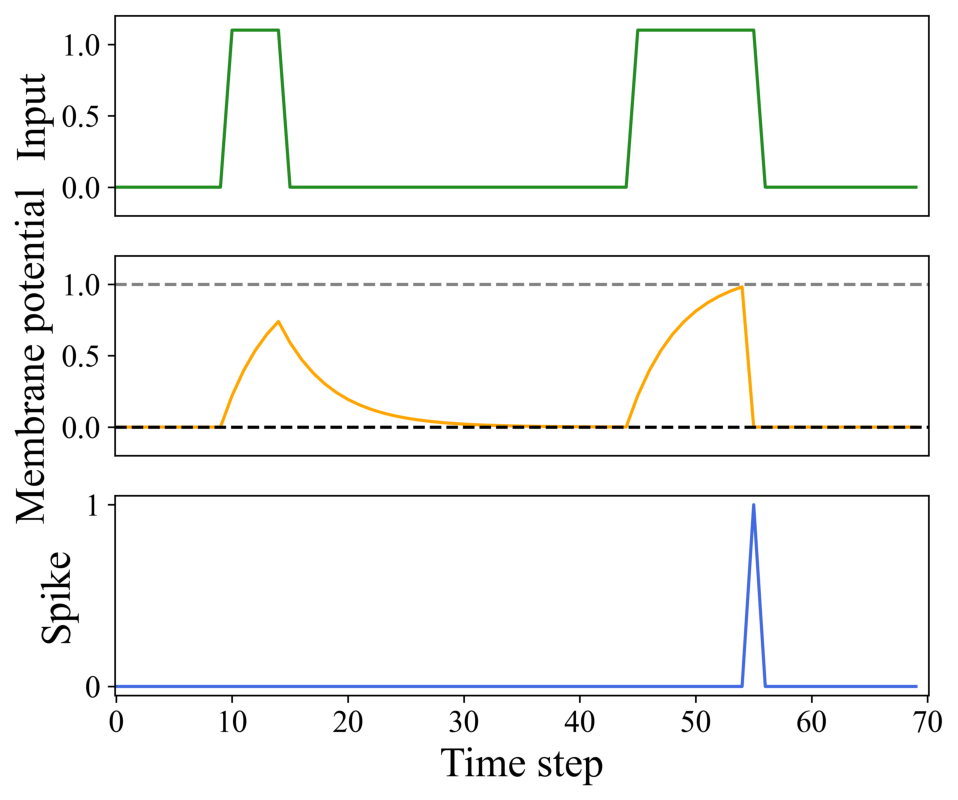}}
    \caption{The LIF neuron. (a) Structure of the LIF neuron. (b) The membrane potential of the LIF neuron rises with the current input. Without input, it decays back to the reset potential. When the membrane potential exceeds the threshold potential, the neuron fires and resets the membrane potential.}     \label{fig_LIF}
\end{figure}

A LIF neuron, similar to the structure of Fang \emph{et al.} \cite{fang2021deep,fang2021incorporating}, is depicted in Figure~\ref{fig_LIF}(a). Its sub-threshold membrane potential dynamic, illustrated in Figure~\ref{fig_LIF}(b), is
\begin{align}
    \tau\frac{dv(t)}{dt} = -(v(t)-v_{\mathrm{reset}}) + x(t), \label{eq_LIF}
\end{align}
where $v(t)$ is the membrane potential, $x(t)$ the input synaptic current, $\tau$ the membrane time constant, and $v_{\mathrm{reset}}$ the reset potential.

The corresponding discrete version is more useful in numerical simulations. The firing process of LIF neurons in layer $l$ is described as:
\begin{enumerate}
\item \emph{Compute the input:}
    \begin{align}
        \bm{x}_t = \bm{W}^{l-1} \bm{s}_{t}^{l-1},  \label{eq_LIF_1}
    \end{align}
    where $\bm{x}_t \in \mathbb{R}^{M^l}$ is the vector of input synaptic currents at time step $t$, in which $t=1,2,\dots,T$ is the time step index ($T$ is the number of forward propagation steps in an iteration), and $M^l$ is the number of neurons in layer $l$. $\bm{s}_{t}^{l-1} \in \{0,1\}^{M^{l-1}}$ are the spikes of neurons from the previous layer and $\bm{W}^{l-1} \in \mathbb{R}^{M^l \times M^{l-1}}$ an input weight matrix. For simplicity, we omit the superscripts $l$ and $M^l$ of all neuronal states in layer $l$, and all equations ignore the bias.

    The Cuba-LIF neuron \cite{dampfhoffer2022investigating,perez2021neural} is more biologically plausible and complex, by introducing temporal dynamics of the input synaptic currents:
    \begin{align}
        \bm{x}_t = \alpha \bm{x}_{t-1} + \bm{W}^{l-1} \bm{s}_{t}^{l-1} + \bm{U}^{l} \bm{s}_{t-1},        \label{eq_Cuba_LIF_1}
    \end{align}
    where $\alpha \in (0,1)$ is the synaptic leaky factor, $\bm{s}_{t-1}$ the spikes at the previous time step, and $\bm{U}^{l}$ the recurrent weight matrix. $\bm{x}_t$ decays $\bm{x}_{t-1}$ at the previous time step, so the synaptic currents are recurrent for the Cuba-LIF neuron.

\item \emph{Leaky and integrate:}    \\
    Consider the forward finite difference formula for (\ref{eq_LIF}):
    \begin{align}
        \bm{h}_t = \beta \bm{v}_t + \left(1-\beta \right) \bm{x}_t, \label{eq_LIF_2}
    \end{align}
    where $\bm{h}_t$ are the hidden membrane potentials, $\bm{v}_t$ are the membrane potentials ($\bm{v}_{\mathrm{reset}}=\bm{0}$ is omitted), and $\beta\in(0,1)$ is the membrane leaky factor:
    \begin{align}
        \beta = 1-\frac{1}{\tau}. \label{eq_LIF_3}
    \end{align}
    Note that $\tau>1$ is required for the approximation to be valid \cite{fang2021incorporating}. For integrate-and-fire (IF) neuron, there is no membrane potential leakage, and equation (\ref{eq_LIF_2}) becomes
    \begin{align}
      \bm{h}_t =\bm{v}_t + \bm{x}_t. \label{eq_IF}
    \end{align}

\item \emph{Fire or not:}
    \begin{align}
        \bm{s}_t = \epsilon \left(\bm{h}_t - \bm{v}_{\mathrm{th}} \right), \label{eq_LIF_4}
    \end{align}
    where $\epsilon (\cdot)$ is the Heaviside activation, and $\bm{v}_{\mathrm{th}}$ the firing thresholds. The neuron fires a spike when the hidden membrane potential exceeds the firing threshold. When performing backpropagation, $\epsilon (\cdot)$ is replaced by the continuous arc tangent function:
    \begin{align}
        \epsilon(x) = \frac{1}{\pi} \arctan (\pi x) + \frac{1}{2}. \label{eq_LIF_5}
    \end{align}

    \item \emph{Reset or not:}
    \begin{align}
    \bm{v}_{t+1} = \left( \bm{1} - \bm{s}_t \right) \odot \bm{h}_t + \bm{s}_t \odot \bm{v}_{\mathrm{reset}}, \label{eq_LIF_6}
    \end{align}
    where $\bm{v}_{\mathrm{reset}}$ are the reset potentials, and $\odot$ is the element-wise multiplication. When the neuron fires, the membrane potential resets.
\end{enumerate}

\subsection{Vanishing gradients of the LIF network} \label{sect:vanish}

Ponghiran and Roy \cite{ponghiran2022spiking} found that the gradients mainly flow along the synaptic current pathway in the Cuba-LIF network. A brief analysis of the LIF network is performed below.

\begin{figure}[htpb]    \centering
    \includegraphics[width=\linewidth]{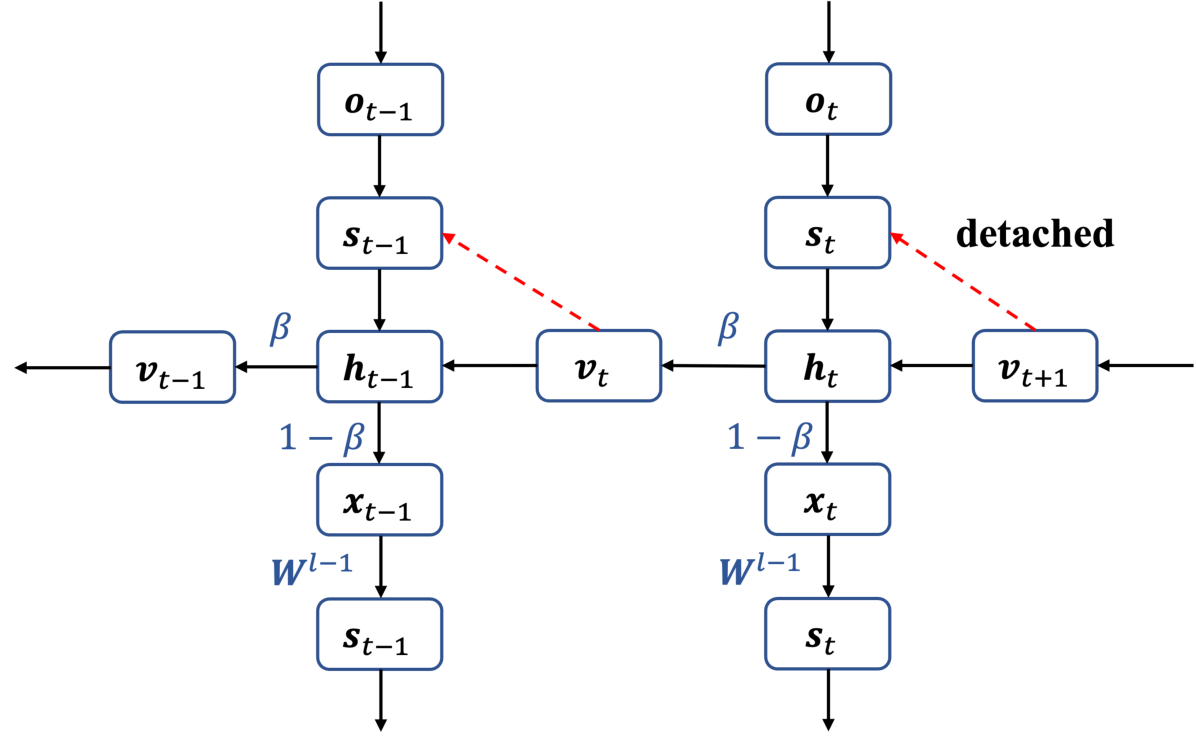}
    \caption{The unrolled computational graph of the LIF network. The red dotted lines represent the detached backward pathways.} \label{fig_BPTT}
\end{figure}

Figure~\ref{fig_BPTT} depicts the unrolled computational graph of a LIF network. Note that we detach spikes $\bm{s}_t$ of (\ref{eq_LIF_6}) during backpropagation because it improves the model performance \cite{fang2021incorporating,zenke2021remarkable} and provides a clearer explanation of the vanishing gradient problem. Therefore, backpropagation does not include the backward path from $\bm{v}_{t+1}$ to $\bm{s}_t$ (the red dotted lines in Figure~\ref{fig_BPTT}). The simplified computational graph is similar to that of the vanilla RNN \cite{pascanu2013difficulty}.

Consider the gradient $\frac{\partial{\bm{o}_t}}{\partial{\bm{W}^{l-1}}}$, where $\bm{o}_t$ is the output at time step $t$ and $\bm{W}^{l-1}$ is the input weight matrix. The gradient $\frac{\partial{\bm{o}_t}}{\partial{\bm{W}^{l-1}}}$ is
\begin{align}
    \frac{\partial{\bm{o}_t}}{\partial{\bm{W}^{l-1}}} = \frac{\partial{\bm{o}_t}}{\partial{\bm{h}_t}} \sum_{1 \leq i \leq t} \left[\left( \prod_{t \geq k > i} \frac{\partial{\bm{h}_k}}{\partial{\bm{h}_{k-1}}} \right) \frac{\partial{\bm{h}_i}}{\partial{\bm{x}_i}} \frac{\partial{\bm{x}_i}}{\partial{\bm{W}^{l-1}}}\right]. \label{eq_bp_1}
\end{align}
For the convenience of explanation, we analyze the individual $j$-th neuron, where the $\prod_{t \geq k > i} \frac{\partial{h_k^j}}{\partial{h_{k-1}^j}}$ is
\begin{align}
    \prod_{t \geq k > i} \frac{\partial{h_k^j}}{\partial{h_{k-1}^j}} &= \prod_{t \geq k > i} \left( \frac{\partial{h_k^j}}{\partial{v_k^j}} \frac{\partial{v_k^j}}{\partial{h_{k-1}^j}} \right)\nonumber\\
     &= \prod_{t \geq k > i} \beta (1 - s_{k-1}^j), \label{eq_bp_2}
\end{align}
in which $\beta\in(0,1)$ is the membrane leaky factor, and an empty $\prod$ is defined to be one. Here $ \prod_{t \geq k > i} \frac{\partial{h_k^j}}{\partial{h_{k-1}^j}} = 0$ when $s_{k-1}^j=1$ for LIF neuron, which causes vanishing gradients.

The IF neuron does not have the membrane leaky factor $\beta$. Other analyses are similar to LIF's. If $\bm{s}_t$ is not detached or the recurrent structure with synaptic current is considered (Cuba-LIF), the computational graph becomes more complex \cite{ponghiran2022spiking}.

\subsection{Gated parametric neuron (GPN)}

As depicted in Figure~\ref{fig_model}, our proposed GPN has four gating functions, each denoted by a capital letter with a tilde: a forget gate $\widetilde{\bm{F}}_t$, an input gate $\widetilde{\bm{I}}_t$, a threshold gate $\widetilde{\bm{T}}_t$ and a bypass gate $\widetilde{\bm{B}}_t$. More specifically, the following calculations, discrete in time, are performed in the GPN:

\begin{figure*}[htpb]    \centering
    \includegraphics[width=\linewidth,clip]{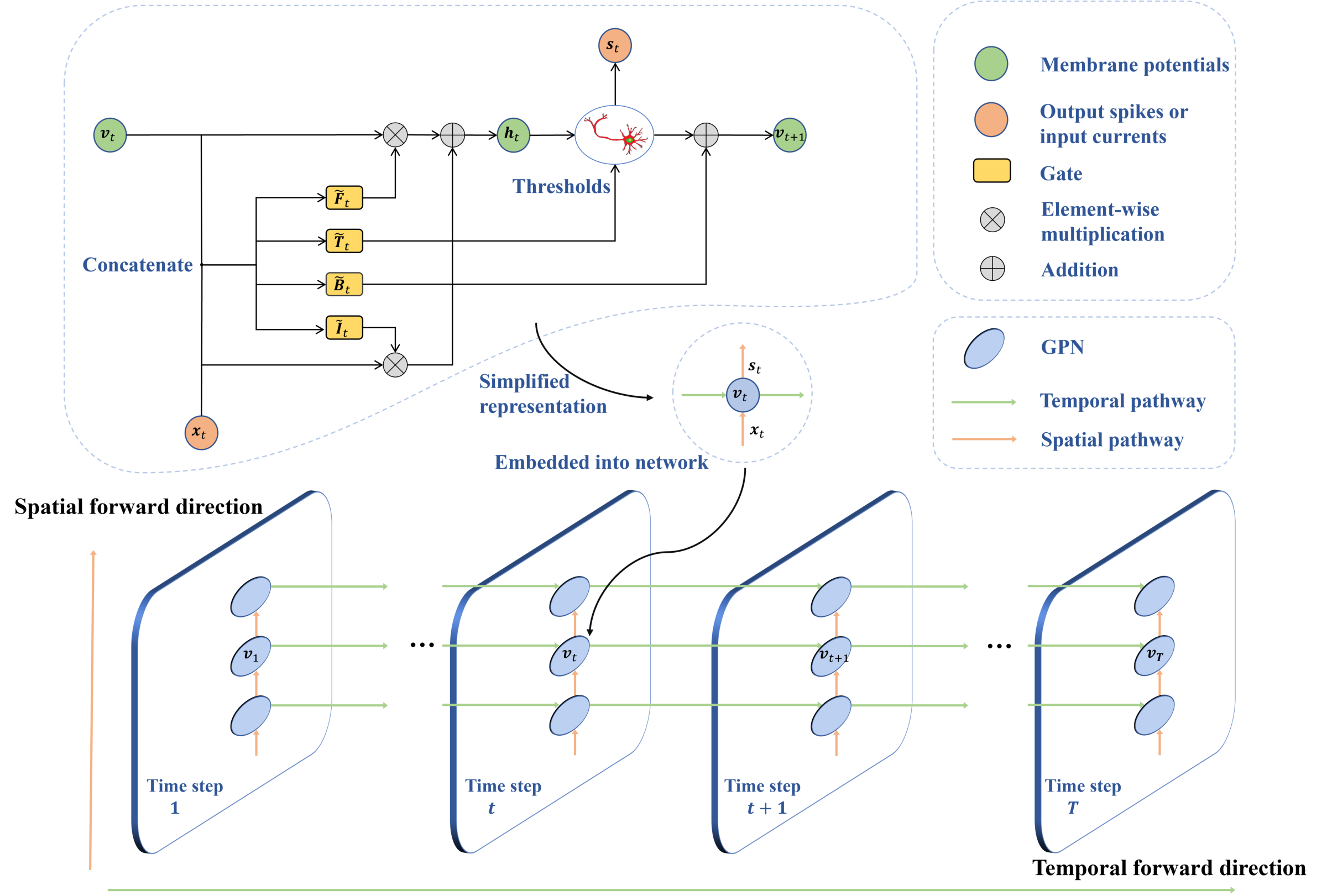}
    \caption{Details of the proposed GPN and the structure of the GPN network. The upper left corner shows the internal structure of GPN, where the green circles represent the membrane potential states, the orange circles the input currents and output spikes, and the yellow boxes the four gate structures of GPN. The spatio-temporal structure of the GPN network is shown at the bottom.}\label{fig_model}
\end{figure*}

\begin{figure}[htpb]      \centering
    \subfigure[]{\includegraphics[width=.45\textwidth,clip]{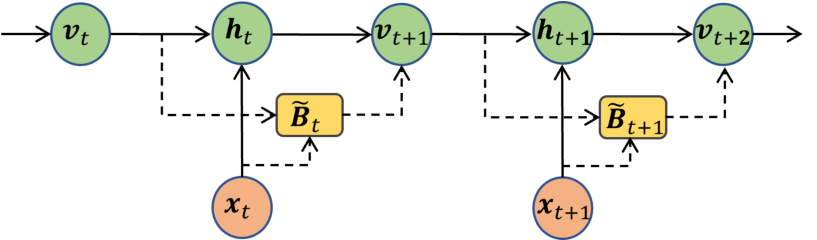}}
    \subfigure[]{\includegraphics[width=.45\textwidth,clip]{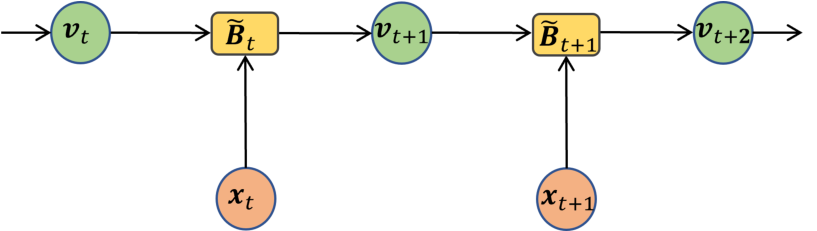}}
    \caption{Structures with the bypass gate. (a) A LIF network with bypass gate; and, (b) a network similar to the vanilla RNN.}     \label{fig_GPN_B}
\end{figure}

\begin{enumerate}
\item \emph{Leaky and integrate:}

    Inspired by the effectiveness of the gating mechanism in mitigating vanishing gradients \cite{Hochreiter1997LSTM,cho2014GRU}, we propose the following empirical gate structure:
    \begin{align}
    \widetilde{\bm{F}}_t &= \sigma \left(\bm{W}_F^v \bm{v}_t + \bm{W}_F^x \bm{x}_t\right), \label{eq_F} \\
    \widetilde{\bm{I}}_t &= \sigma \left(\bm{W}_I^v \bm{v}_t + \bm{W}_I^x \bm{x}_t\right), \label{eq_I} \\
    \bm{h}_t &= \widetilde{\bm{F}}_t \odot \bm{v}_t + \widetilde{\bm{I}}_t \odot \bm{x}_t, \label{eq_GPN_1}
    \end{align}
    where $\widetilde{\bm{F}}_t$ is the forget gate and $\widetilde{\bm{I}}_t$ the input gate. $\sigma$ is the sigmoid function, and $\bm{W}_F^v$, $\bm{W}_F^x$, $\bm{W}_I^v$ and $\bm{W}_I^x$ are weight matrices of the gates. Membrane potentials $\bm{v}_t$ and synaptic inputs $\bm{x}_t$ at each time step $t$ are used as the gate inputs. $\bm{h}_t$ are the hidden membrane potentials. $\widetilde{\bm{F}}_t$ and $\widetilde{\bm{I}}_t$ replace the membrane leaky factors $\beta$ and $1-\beta$ in (\ref{eq_LIF_2}). Note that replacing $\beta \in (0,1)$ with $\widetilde{\bm{F}}_t$ does not change the range of values, and the same applies to $\widetilde{\bm{I}}_t$.

    The gradient on the membrane potential pathway is jointly calculated by the forget gate $\widetilde{\bm{F}}_t$ and the input gate $\widetilde{\bm{I}}_t$.

\item \emph{Fire or not:}

    The heterogeneous thresholds are generated by:
    \begin{align}
    \widetilde{\bm{T}}_t &= \sigma \left(\bm{W}_T^v \bm{v}_t + \bm{W}_T^x \bm{x}_t\right), \label{eq_T} \\
    \bm{s}_t &= \epsilon \left(\bm{h}_t - \widetilde{\bm{T}}_t\right), \label{eq_GPN_2}
    \end{align}
    where $\bm{W}_T^v$ and $\bm{W}_T^x$ are weight matrices of the threshold gate $\widetilde{\bm{T}}_t$. The gate is essentially a linear layer with activation, so its output values are different between neurons and determined by the inputs and the weight matrix. Specifically, $\widetilde{\bm{F}}_t$, $\widetilde{\bm{I}}_t$ and $\widetilde{\bm{T}}_t$ represent the membrane leaky factors and thresholds of a layer of neurons at time step $t$. Different neurons have different weights, so the leaky factors and thresholds are spatially heterogeneous. Since the gates are computed with $\bm{v}_t$ and $\bm{x}_t$ dynamic in time, the leaky factors and thresholds are also temporally heterogeneous. In addition, setting initial values for the neuronal parameters is no longer needed, because GPN uses membrane potentials and synaptic currents to compute them.

\item \emph{Reset or not:}

    Since the gate input contains membrane potentials $\bm{v}_t$, we design a bypass gate and add the gate output to $\bm{v}_{t+1}$ at the next time step. We add the bypass gate in parallel to the membrane potential pathway so a hybrid RNN-SNN structure is formed without destroying the spiking dynamics:
    \begin{align}
    \widetilde{\bm{B}}_t &= \sigma \left(\bm{W}_B^v \bm{v}_t + \bm{W}_B^x \bm{x}_t\right), \label{eq_B} \\
    \bm{v}_{t+1} &= \left(\bm{1}-\bm{s}_t \right) \odot \bm{h}_t + \bm{s}_t \odot \bm{v}_{\mathrm{reset}} + \widetilde{\bm{B}}_t, \label{eq_GPN_3}
    \end{align}
    where $\widetilde{\bm{B}}_t$ is the bypass gate, and $\bm{W}_B^v$ and $\bm{W}_B^x$ are the corresponding weight matrices. The gate structure is essentially a linear layer with activation. If the gate input and output are treated as hidden states, it is similar to RNN. When considering only LIF with $\widetilde{\bm{B}}_t$, it has a hybrid RNN-SNN structure, as shown in Figure~\ref{fig_GPN_B}(a). Specifically, Figure~\ref{fig_GPN_B}(b) shows that if we ignore the neuronal terms containing $\bm{s}_t$ and $\bm{h}_t$ in (\ref{eq_GPN_3}), then the structure is similar to the vanilla RNN.
\end{enumerate}

\subsection{Comparison with gated networks}

\begin{enumerate}
    \item \emph{LSTM/GRU}. Unlike LSTM with two hidden states, GPN only contains one state of membrane potential, closer to the structure of GRU. For neuron outputs, GPN uses the spiking mechanism to fire spikes and reset membrane potential, without using the output gates in LSTM and reset gates in GRU. For neuronal dynamics, GPN adds a threshold gate to compute the thresholds and a bypass gate to obtain the hybrid RNN-SNN structure. In summary, GPN is a more sophisticated gated SNN structure.

    \item \emph{Gated SNNs}. Unlike previous work \cite{dampfhoffer2022investigating}, the gate inputs of GPN use the membrane potentials and synaptic currents instead of the spikes, making it more similar to the original GRU. More importantly, the gradient mainly flows along the synaptic current pathway for the Cuba-LIF network \cite{ponghiran2022spiking}, so modifying the gradient of the membrane potential pathway only is insufficient \cite{dampfhoffer2022investigating}. Since the LIF network has a single temporal membrane potential pathway, the modification of GPN maintains more gradient information. Ponghiran and Roy \cite{ponghiran2022spiking} modified the Cuba-LIF network with gates, whereas we also focus on the heterogeneous neuronal parameters, in addition to the vanishing gradient problem. GLIF \cite{yao2022glif} utilized gating mechanisms to fuse multiple biological features, including membrane potential leakage, integration accumulation, and spike initiation. Unlike GLIF, GPN focuses on the vanishing gradient problem. The experimental results of the relevant approaches are presented in the following text.

\end{enumerate}

\section{Experimental Settings} \label{sect:Experiments}

\subsection{Datasets}

\begin{figure}[htpb]      \centering
    \subfigure[]{\includegraphics[width=.24\textwidth,clip]{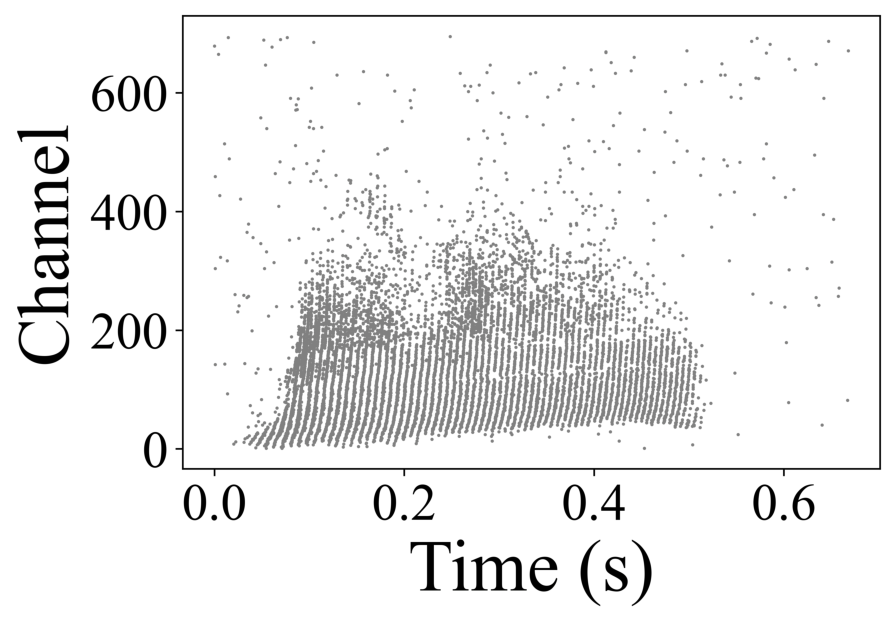}}
    \subfigure[]{\includegraphics[width=.24\textwidth,clip]{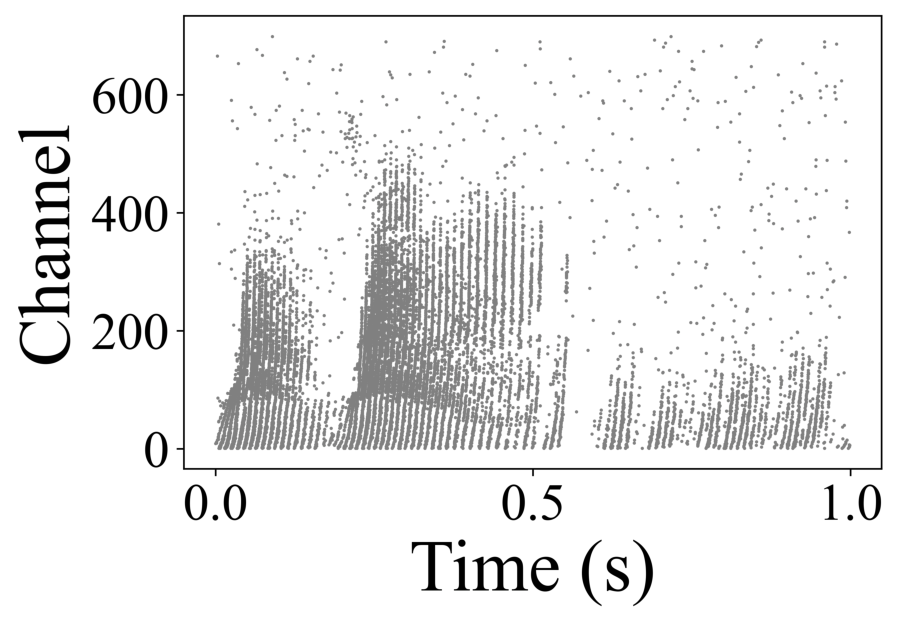}}
    \caption{The two spike-based audio datasets. (a) A sample from the SHD dataset; and, (b) a sample from the SSC dataset.} \label{fig_datasets}
\end{figure}

Spiking Heidelberg digits (SHD) and spiking speech commands (SSC) are public datasets released under the Creative Commons Attribution 4.0 International Licence \cite{cramer2022heidelberg}. SHD is a spike-based audio dataset comprised of high-quality recordings of English and German spoken digits from zero to nine in 20 classes. SHD has 8,156 and 2,264 digital sequences for training and test, respectively. We further reserved 15\% sequences from the training set for validation. SSC is a larger-scale dataset transformed from speech commands v0.2, which contains 35 different classes of spoken words and 105,829 audio files. It has already been split into 75,466, 9,981, and 20,382 samples for training, validation, and test. The spikes of both datasets have 700 input channels. Figure\ref{fig_datasets} depicts the spike-based audio samples from SHD and SSC, respectively. We performed speech recognition tasks on both datasets.

For data preprocessing, audio samples shorter than one second were padded with zeros, and those longer than one second were truncated, as in previous work \cite{yin2020effective}. Then, we binned them into $T$ time steps and counted the spikes in each time step. We randomly rolled the input 0-15 channels at each time step for data augmentation.

\subsection{Implementation details}

SpikingJelly \cite{fang2023spikingjelly} was used to train the SNNs. We employed the Adam optimizer, whose learning rate decayed according to the loss on the training set. We averaged the model outputs across all time steps and applied the cross-entropy loss. Table~\ref{table_params} shows the detailed hyper-parameters.

The SNN structures contain fully connected layers, GPN layers, and dropout layers. In Table~\ref{table_stru}, $FC1024$ is a fully connected layer with 1024 output channels, and $GPN1024$ is a GPN layer with 1024 neurons. $DP$ is a dropout layer specially designed for SNNs \cite{lee2020enabling}. The obtained models were trained with early stopping on the validation set, and all experiments were repeated three times.

\begin{table}[htpb]
    \caption{Hyper-parameter settings.} \label{table_params}
    \center
    \renewcommand\arraystretch{1.3}
    \setlength{\tabcolsep}{14pt}{
    \begin{tabular}{ccc}
    \toprule
    \textbf{Hyper-parameter} & \textbf{SHD} & \textbf{SSC} \\
    \midrule
    $T$ & 40 & 60\\
    Learning rate & 5e-4 & 1e-3 \\
    Batch size & 128 & 1024\\
    Max number of epochs & 150 & 300\\

    \bottomrule
    \end{tabular}}
\end{table}

\begin{table}[htpb]
    \caption{Network structures.} \label{table_stru}
    \center
    \renewcommand\arraystretch{1.3}
    \setlength{\tabcolsep}{3pt}{
    \begin{tabular}{cc}
    \toprule
    \textbf{Dataset} & \textbf{Network structure}  \\
    \midrule

    SHD & $FC1024-GPN1024-DP-FC20$ \\
    SSC & $(FC1024-GPN1024-DP)*3-FC35$ \\

    \bottomrule
    \end{tabular}}
\end{table}

\section{Results} \label{sect:Results}

\subsection{Classification performance}

Table~\ref{table_sota} shows the performance of different neuron modification approaches on the two datasets. The upper panel shows the results from the original publications, and the lower panel shows the results under our experimental settings.

\begin{table}[htpb]\center
    \caption{Performance comparison of GPN with existing approaches. } \label{table_sota}
    \renewcommand\arraystretch{1.3}
    \setlength{\tabcolsep}{12pt}{
    \begin{tabular}{cccc}        \toprule
        \textbf{Methods} & \textbf{SHD}& \textbf{SSC} \\
        \midrule
            Cuba-LIF \cite{cramer2022heidelberg}  & $83.2 \pm 1.3$ & $50.9\pm1.1$ \\
            Cuba-LIF \cite{perez2021neural} & $82.7 \pm 0.8$  & $60.1 \pm 0.7$ \\
            ALIF \cite{yin2021accurate} & $90.4$ & $74.2$ \\
            SpikGRU \cite{dampfhoffer2022investigating}  & $86.4 \pm 1.8$ &  $77.0 \pm 0.4$ \\
            ANN \cite{cramer2022heidelberg,dampfhoffer2022investigating}  & $\bm{92.4} \pm 0.7$ & $78.1 \pm 0.3$  \\
        \hline
            PLIF \cite{fang2021incorporating} & $71.7 \pm 0.4$ & $70.4 \pm 0.2$ \\
            IF & $79.0 \pm0.3 $ & $46.6 \pm 0.1$ \\
            LIF & $75.8 \pm 0.2$ & $67.4 \pm 0.4$ \\
            Cuba-LIF & $78.8 \pm 1.2$ & $72.8 \pm 2.3$ \\
            \textbf{GPN} & $90.8 \pm 0.1$ & $\bm{78.3} \pm 0.3$ \\
     \bottomrule
    \end{tabular}}
\end{table}

Cuba-LIF \cite{cramer2022heidelberg} should be viewed as the performance baseline for SNNs. Cuba-LIF \cite{perez2021neural} constructs heterogeneous time constants on temporally complex auditory tasks. ALIF \cite{yin2021accurate} is a more complex and heterogeneous structure with adaptive firing thresholds and trainable time constants. SpikGRU \cite{dampfhoffer2022investigating} combines Cuba-LIF and GRU. PLIF \cite{fang2021incorporating} uses trainable time constants.

Compared with existing SNN approaches, our proposed GPN achieved the best performance on the SHD dataset, though slightly worse than the state-of-the-art ANN \cite{cramer2022heidelberg}. GPN also significantly outperformed commonly used SNNs (IF, LIF, and Cuba-LIF) under the same experimental settings. On the SSC dataset, GPN achieved the highest accuracy among all approaches, even $0.2\%$ higher than the state-of-the-art ANN \cite{dampfhoffer2022investigating}. These results demonstrated GPN's advantages in spike-based audio classification, and its potential to outperform ANNs.

\subsection{Ablation Studies}

Ablation studies were performed to verify the necessity and effectiveness of different gates. The following models were considered:
\begin{enumerate}
    \item \emph{GPN w/o $\widetilde{\bm{F}}\&\widetilde{\bm{I}}$}:
        This model did not include the forget gate $\widetilde{\bm{F}}$ and the input gate $\widetilde{\bm{I}}$. The membrane leaky factors $\beta$ were determined by grid search.
    \item \emph{GPN w/o $\widetilde{\bm{T}}$}:
        This model did not include the threshold gate $\widetilde{\bm{T}}$. The firing thresholds $v_{\mathrm{th}}$ were determined by grid search.
    \item \emph{GPN w/o $\widetilde{\bm{B}}$}:
        This model did not include the bypass gate $\widetilde{\bm{B}}$.
\end{enumerate}

The test accuracies are shown in Table~\ref{table_abl}. The GPN network with all four gates achieved the highest accuracy on both datasets. On the SHD dataset, the test accuracy degraded by $3.3\%$ without $\widetilde{\bm{F}}$ and $\widetilde{\bm{I}}$, indicating the effectiveness of $\widetilde{\bm{F}}$ and $\widetilde{\bm{I}}$ in mitigating vanishing gradients. The model with $\widetilde{\bm{T}}$ improved the performance by $0.4\%$ and $0.7\%$ on SHD and SSC, respectively, demonstrating the benefits of heterogeneous neuronal parameters. The bypass gate $\widetilde{\bm{B}}$ was more effective on the SSC dataset, resulting in $7.0\%$ accuracy improvement, demonstrating the effectiveness of the hybrid GPN structure.

\begin{table}[htpb]
    \caption{Ablation Studies.} \label{table_abl}     \center
    \renewcommand\arraystretch{1.3}
    \setlength{\tabcolsep}{12pt}{
    \begin{tabular}{ccc}
    \toprule
    \textbf{Model} & \textbf{SHD} & \textbf{SSC} \\
    \midrule
    GPN w/o $\widetilde{\bm{F}}\&\widetilde{\bm{I}}$  & $87.5 \pm 0.3$ & $77.5 \pm 0.2$  \\
    GPN w/o $\widetilde{\bm{T}}$ & $90.4 \pm 0.6$ & $77.6 \pm 1.2$ \\
    GPN w/o $\widetilde{\bm{B}}$ & $89.2 \pm 0.7$ & $71.3 \pm 0.3$  \\
    \hline
    \textbf{GPN} & $\bm{90.8} \pm 0.1$ & $\bm{78.3} \pm 0.3$ \\
    \bottomrule
    \end{tabular}}
\end{table}

\subsection{Vanishing gradients of SNNs}

\begin{figure}[htpb]      \centering
    \subfigure[]{\includegraphics[width=.24\textwidth,clip]{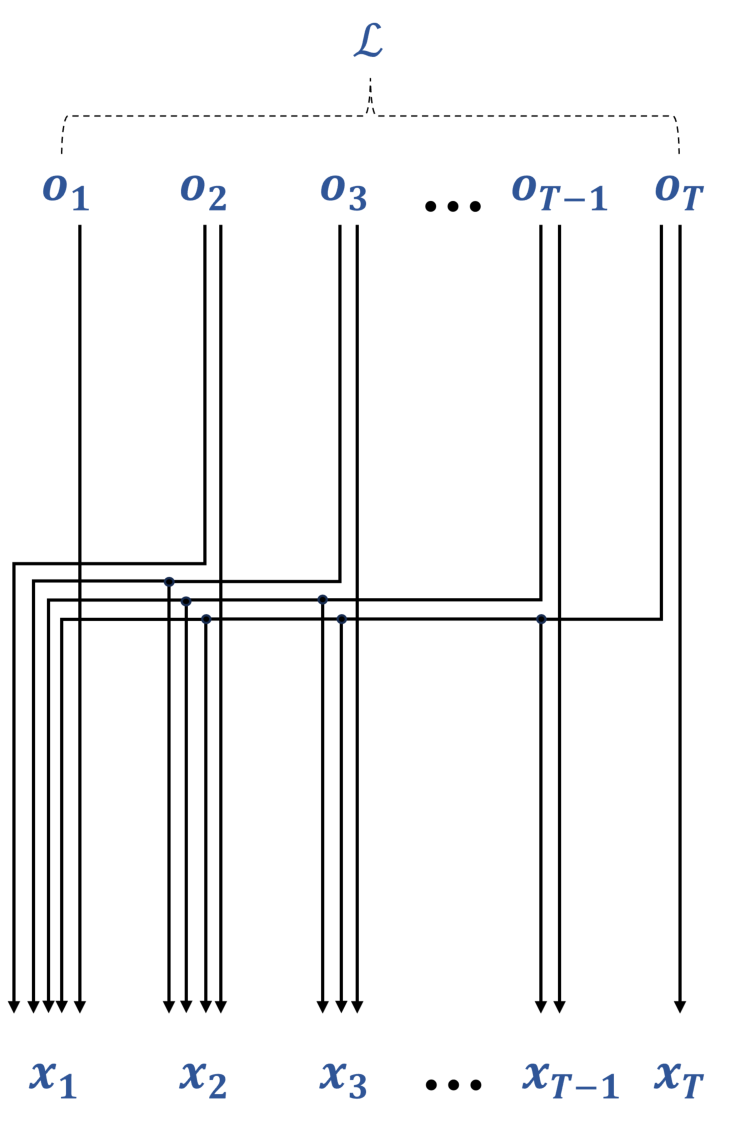}}
    \subfigure[]{\includegraphics[width=.24\textwidth,clip]{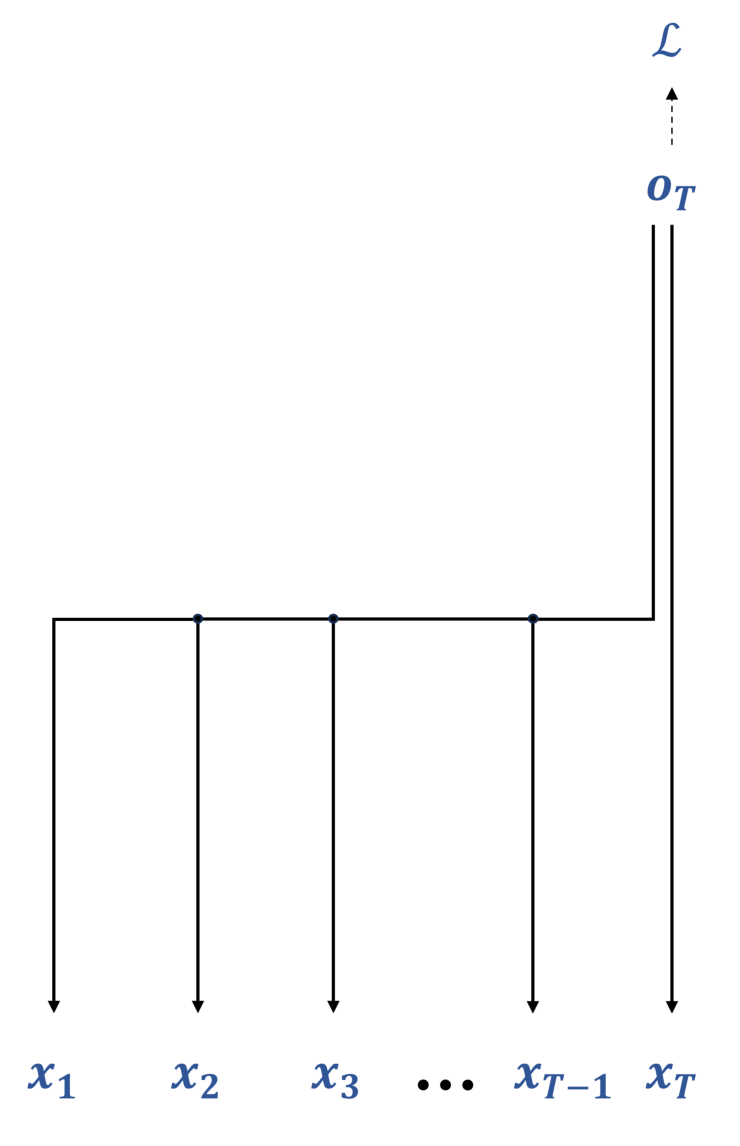}}
    \caption{The backpropagation pathway of two different loss functions. (a) all-step-loss; and, (b) last-step-loss.} \label{fig_loss}
\end{figure}

This subsection analyzes the vanishing gradient problem. We distinguish between two losses, whose computational graphs are depicted in Figure~\ref{fig_loss}. We used \emph{all-step-loss} in previous subsections, but use \emph{last-step-loss} in this subsection to highlight the impact of vanishing gradients.

\begin{enumerate}
\item \emph{All-step-loss}, in which all time step outputs $\bm{o}_i,\bm{o}_{i+1},\dots,\bm{o}_T$ are used to calculate the loss $\sum_{i \leq t \leq T} \frac{\partial{\bm{o}_t}}{\partial{\bm{x}_i}}$. It was used in previous experiments.

\item \emph{Last-step-loss}, in which the last time step output $\bm{o}_T$ is used to compute the loss $\frac{\partial{\bm{o}_T}}{\partial{\bm{x}_i}}$.
\end{enumerate}

\begin{figure*}[htpb]    \centering
  \subfigure{\includegraphics[width=.92\linewidth]{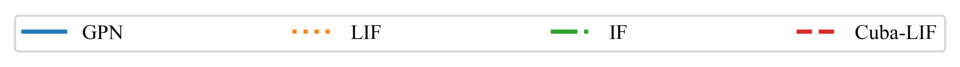}}
  \setcounter{subfigure}{0}
  \subfigure[]{
  \begin{minipage}[b]{0.23\linewidth}
  \includegraphics[width=1\linewidth]{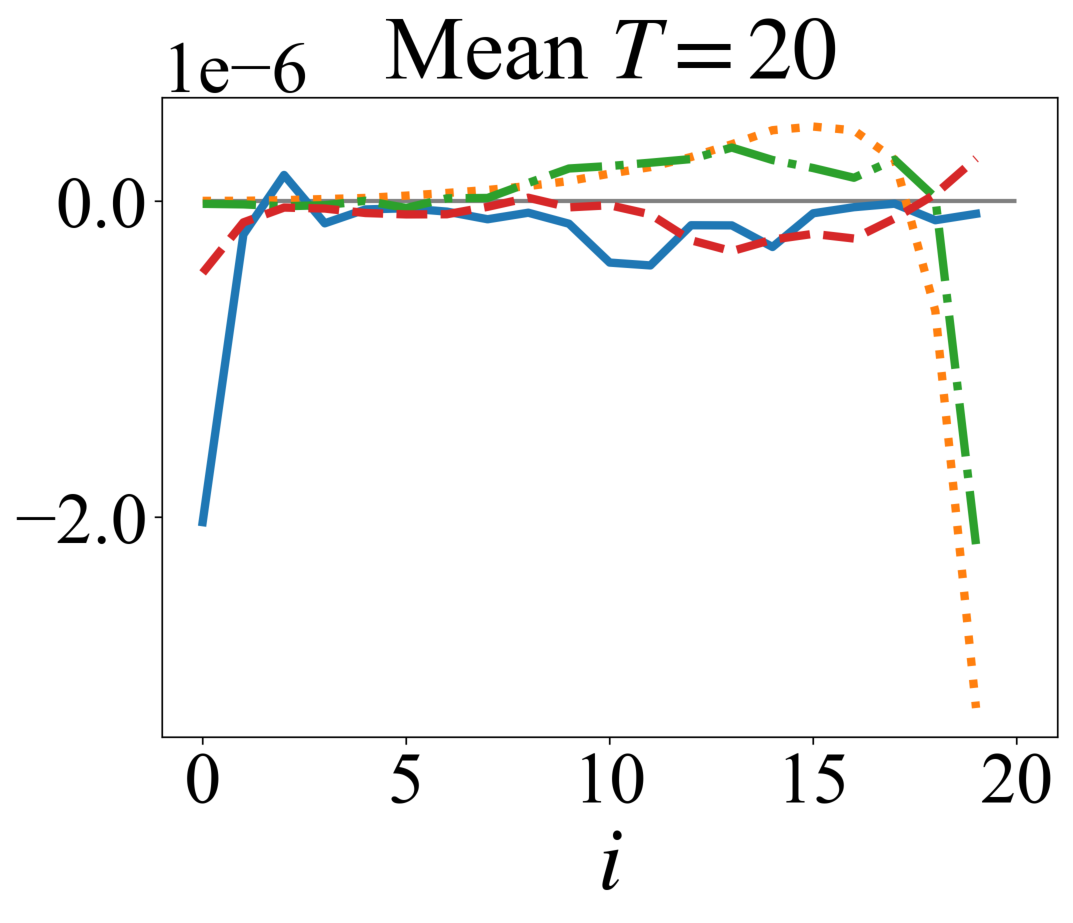}\vspace{4pt}
  \includegraphics[width=1\linewidth]{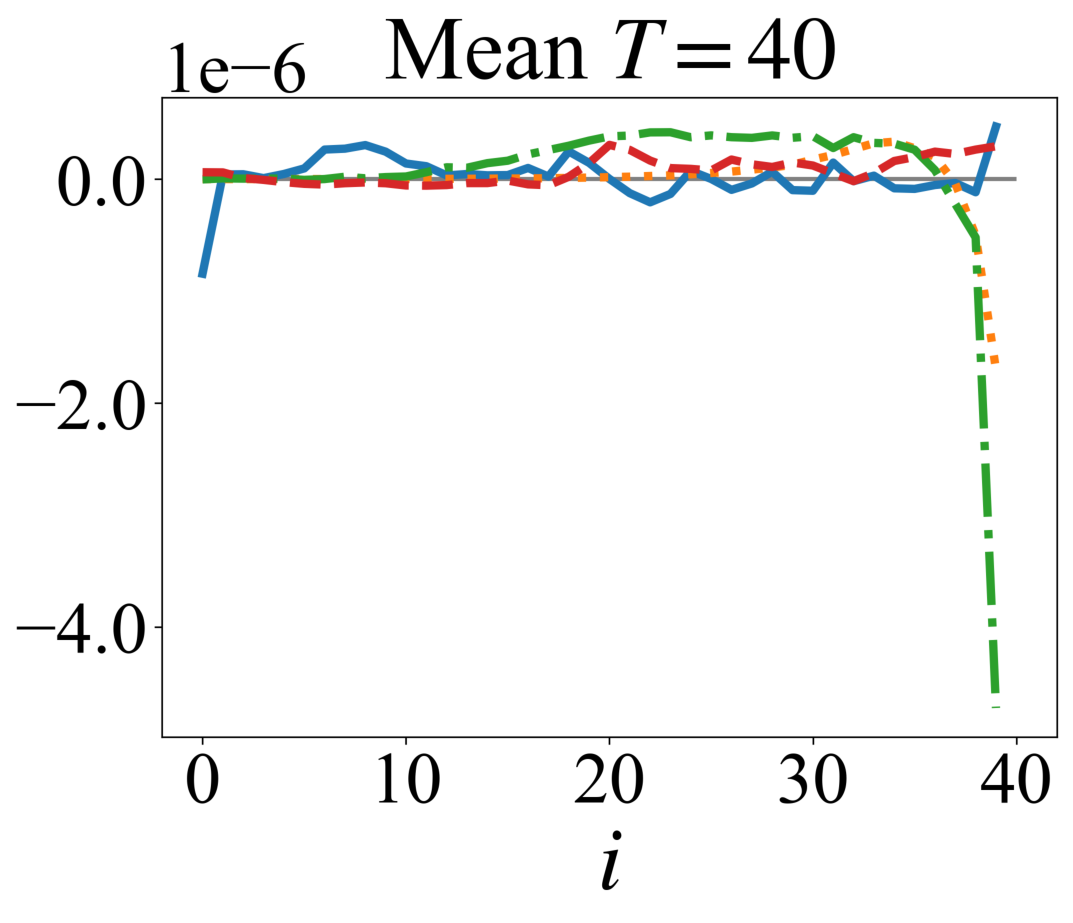}\vspace{4pt}
  \includegraphics[width=1\linewidth]{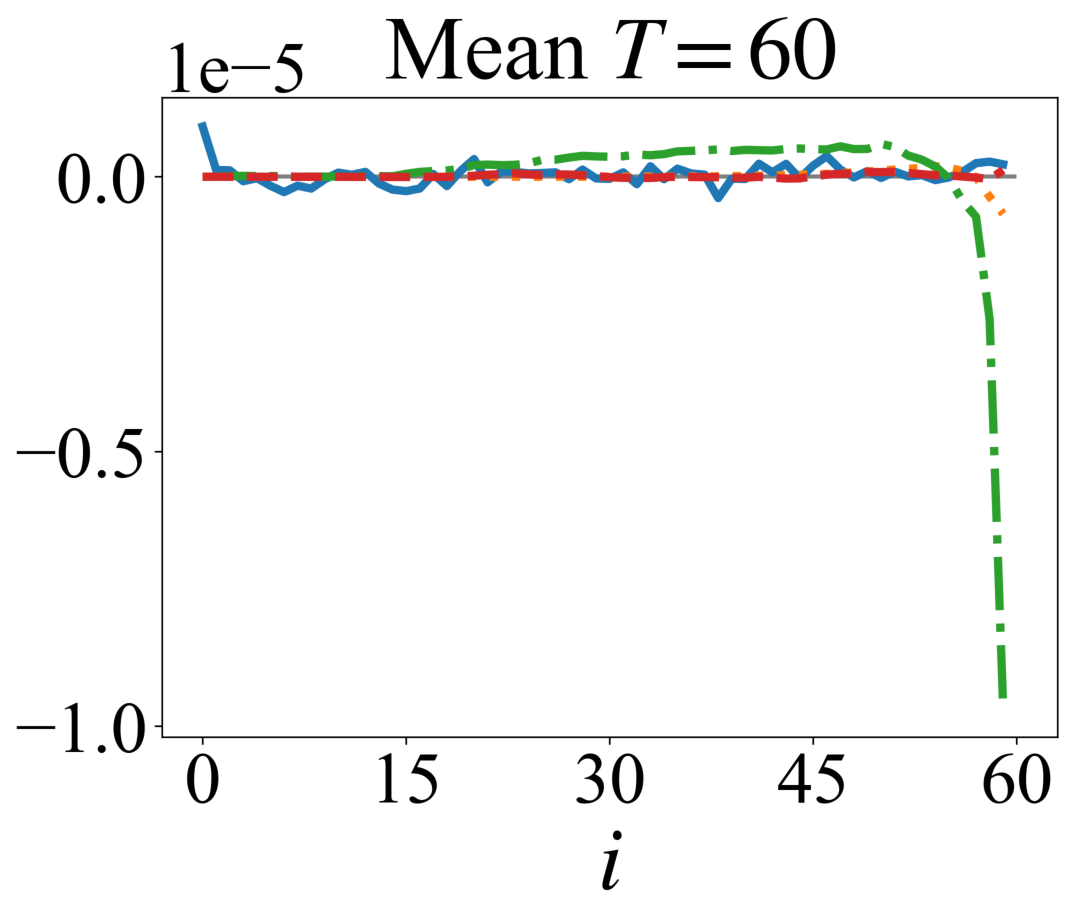}\vspace{4pt}
  \includegraphics[width=1\linewidth]{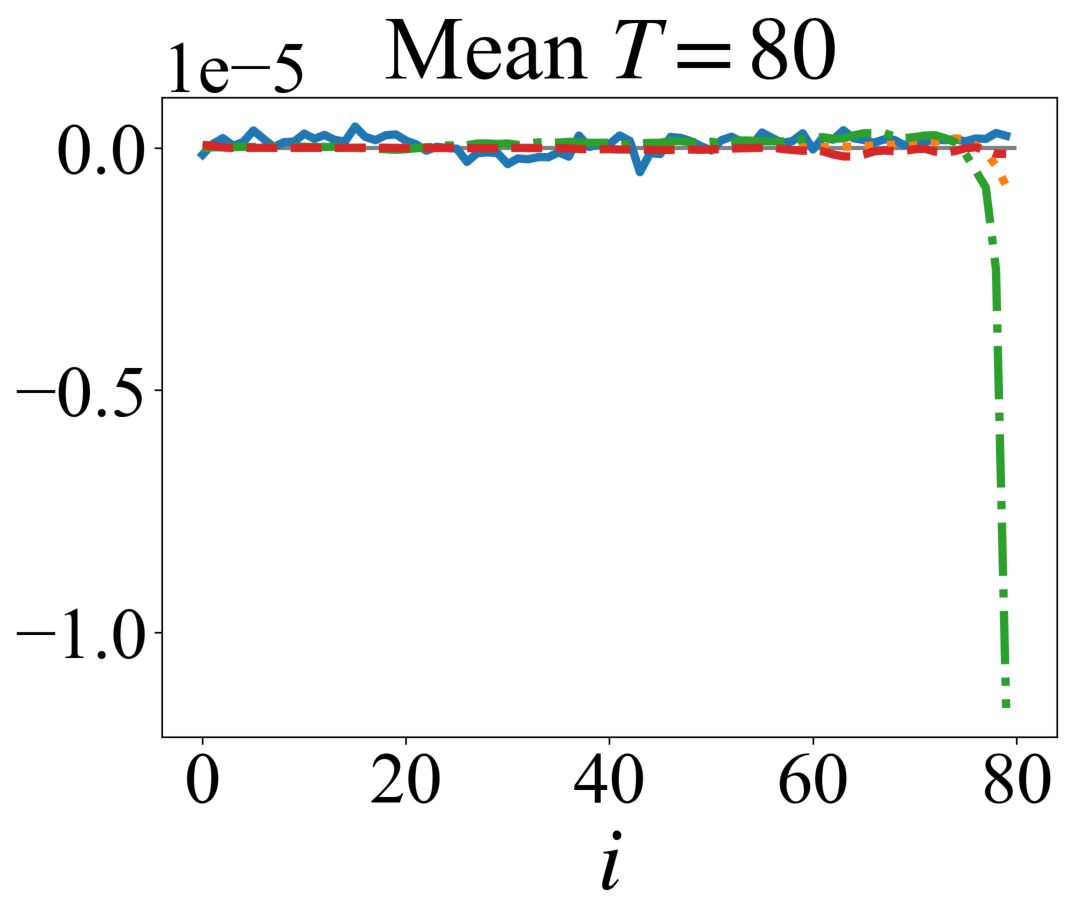}\vspace{4pt}
  \includegraphics[width=1\linewidth]{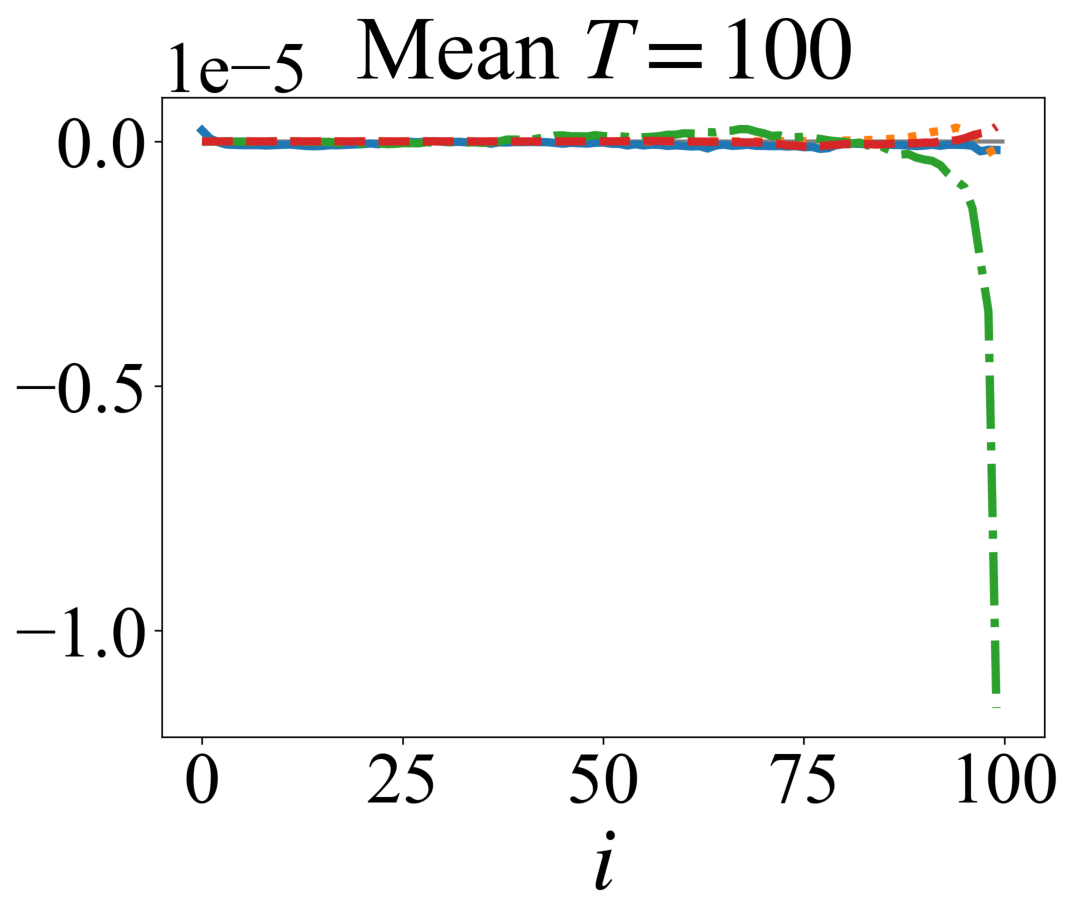}
  \end{minipage}}
  \subfigure[]{
  \begin{minipage}[b]{0.23\linewidth}
  \includegraphics[width=1\linewidth]{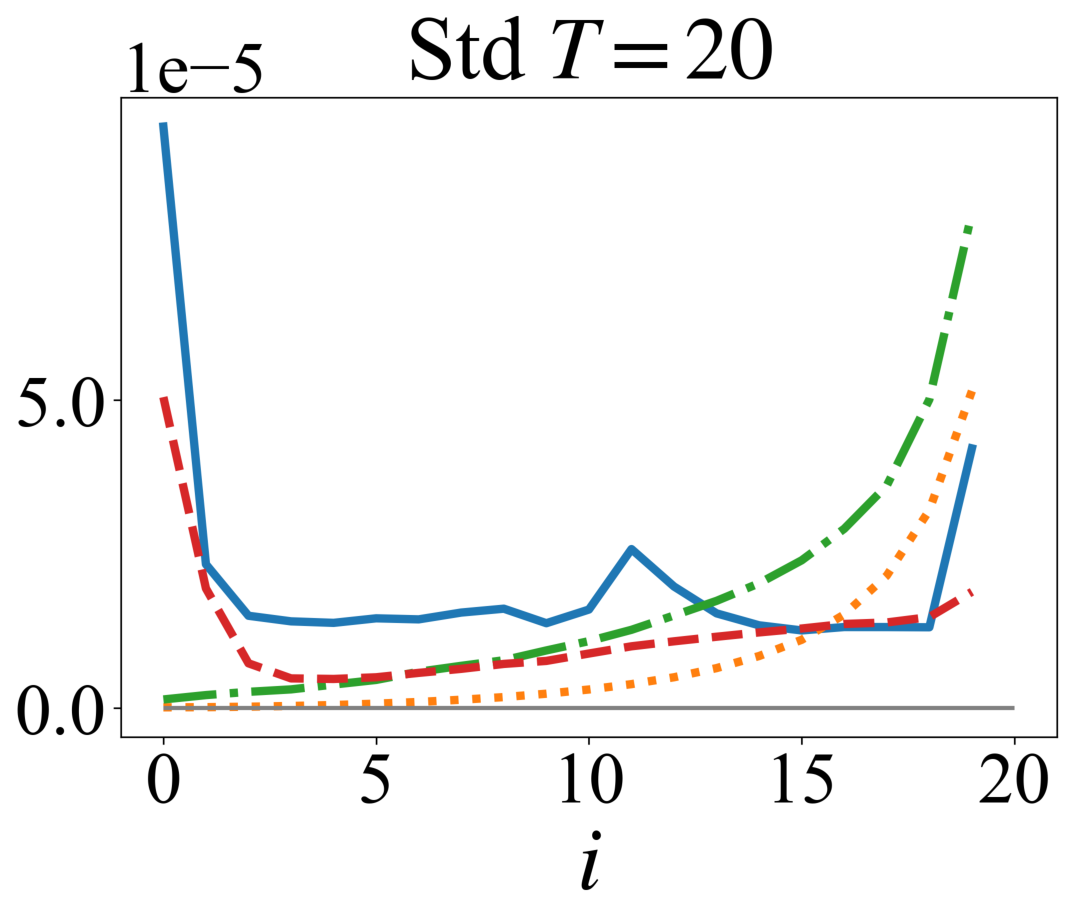}\vspace{4pt}
  \includegraphics[width=1\linewidth]{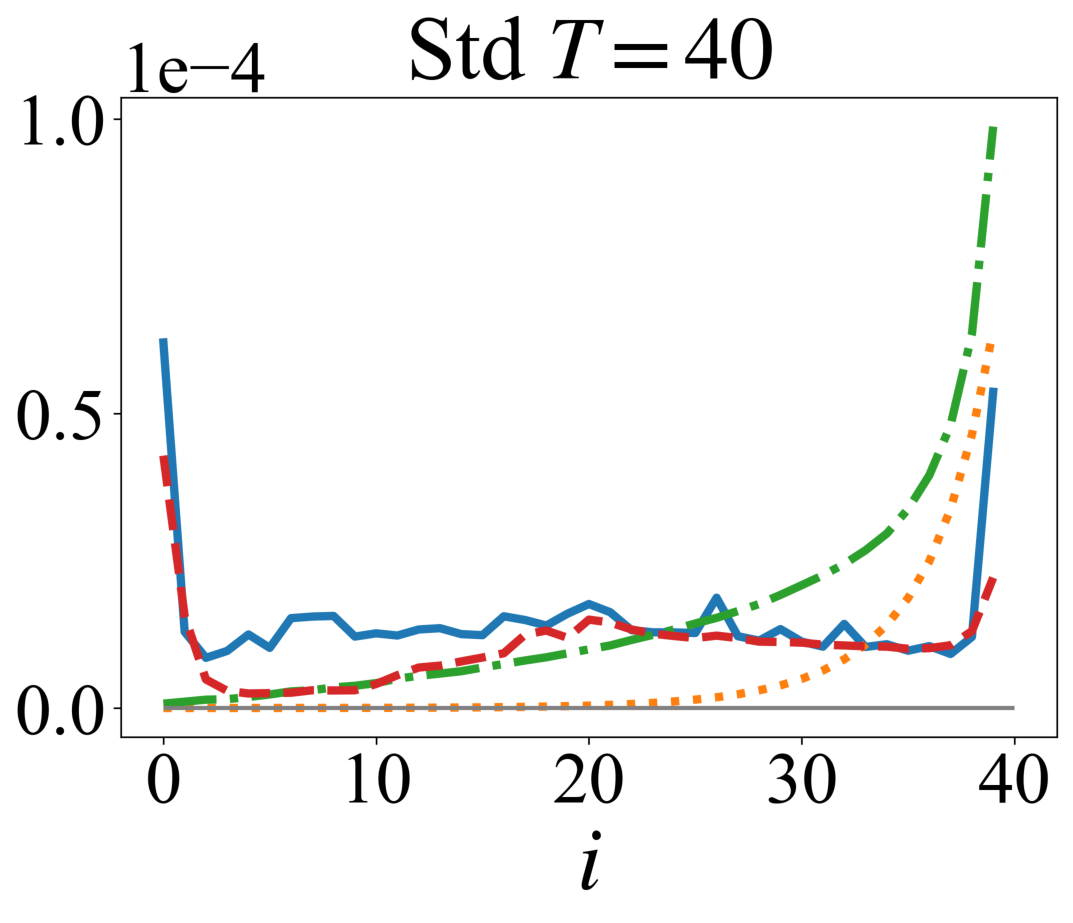}\vspace{4pt}
  \includegraphics[width=1\linewidth]{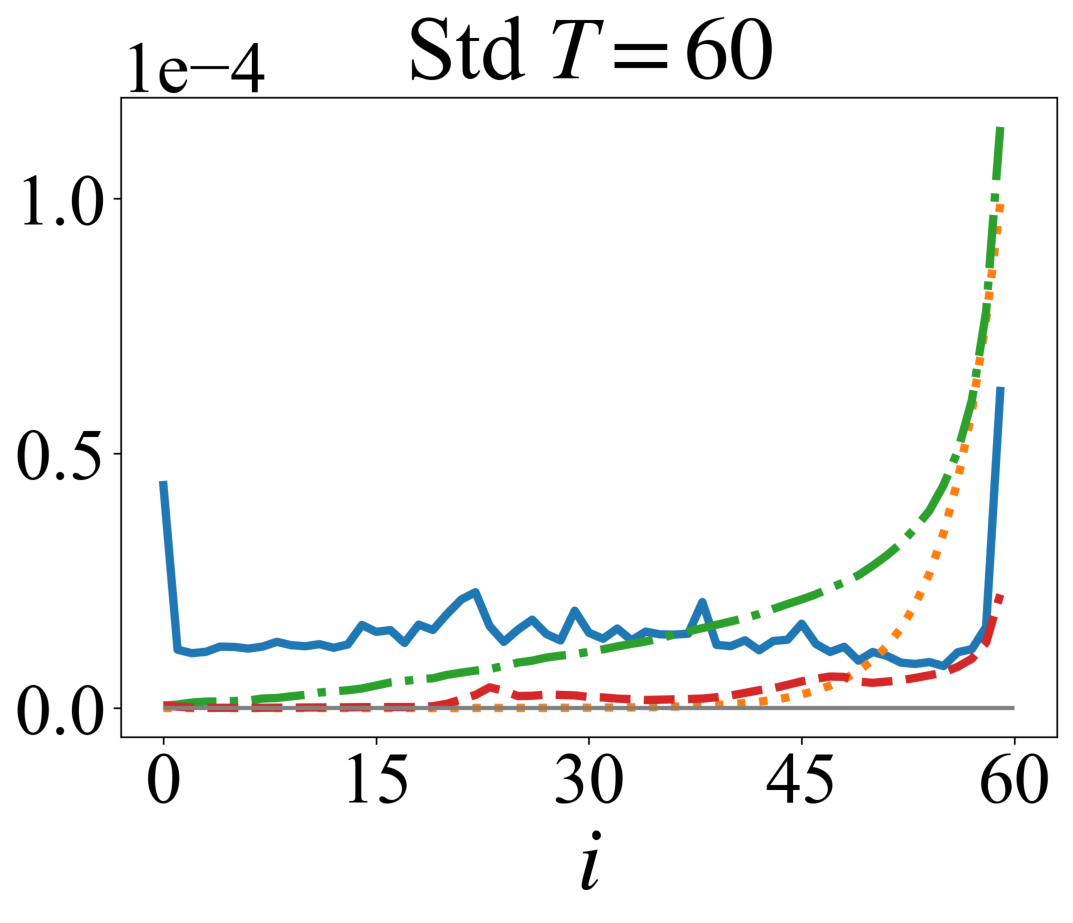}\vspace{4pt}
  \includegraphics[width=1\linewidth]{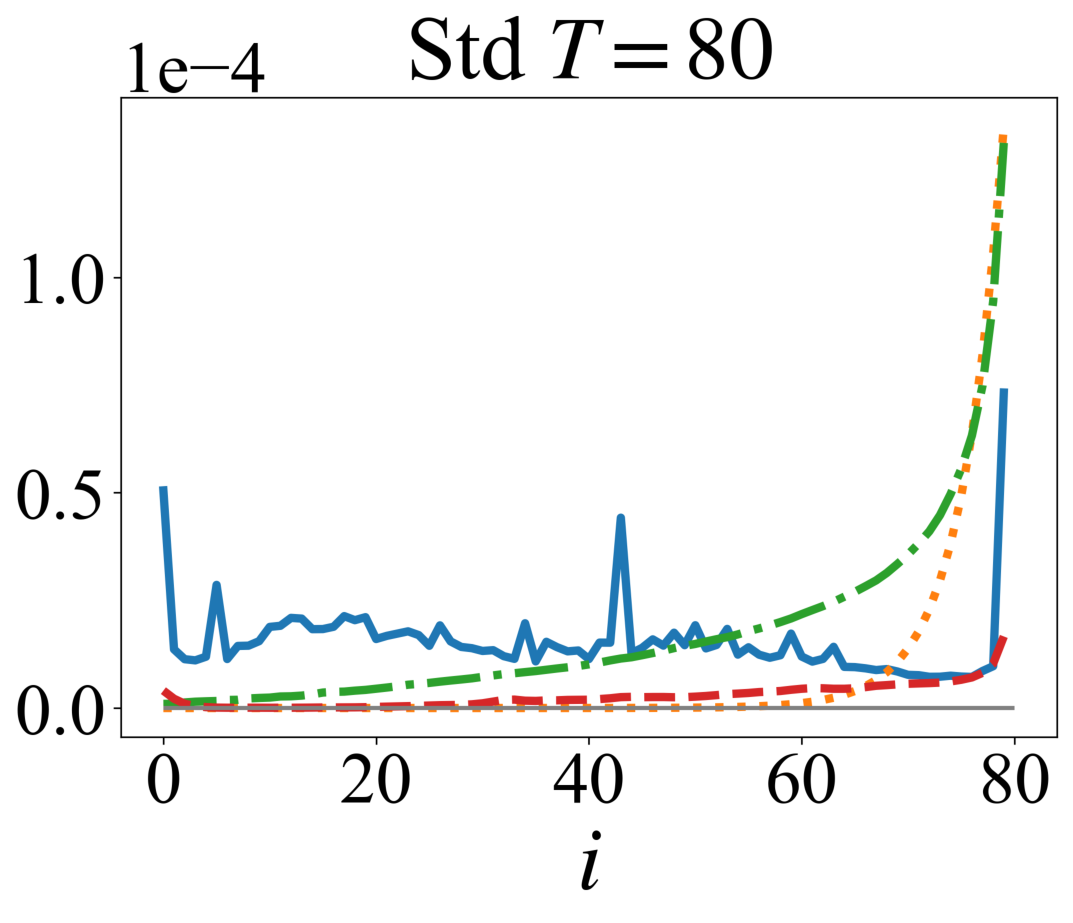}\vspace{4pt}
  \includegraphics[width=1\linewidth]{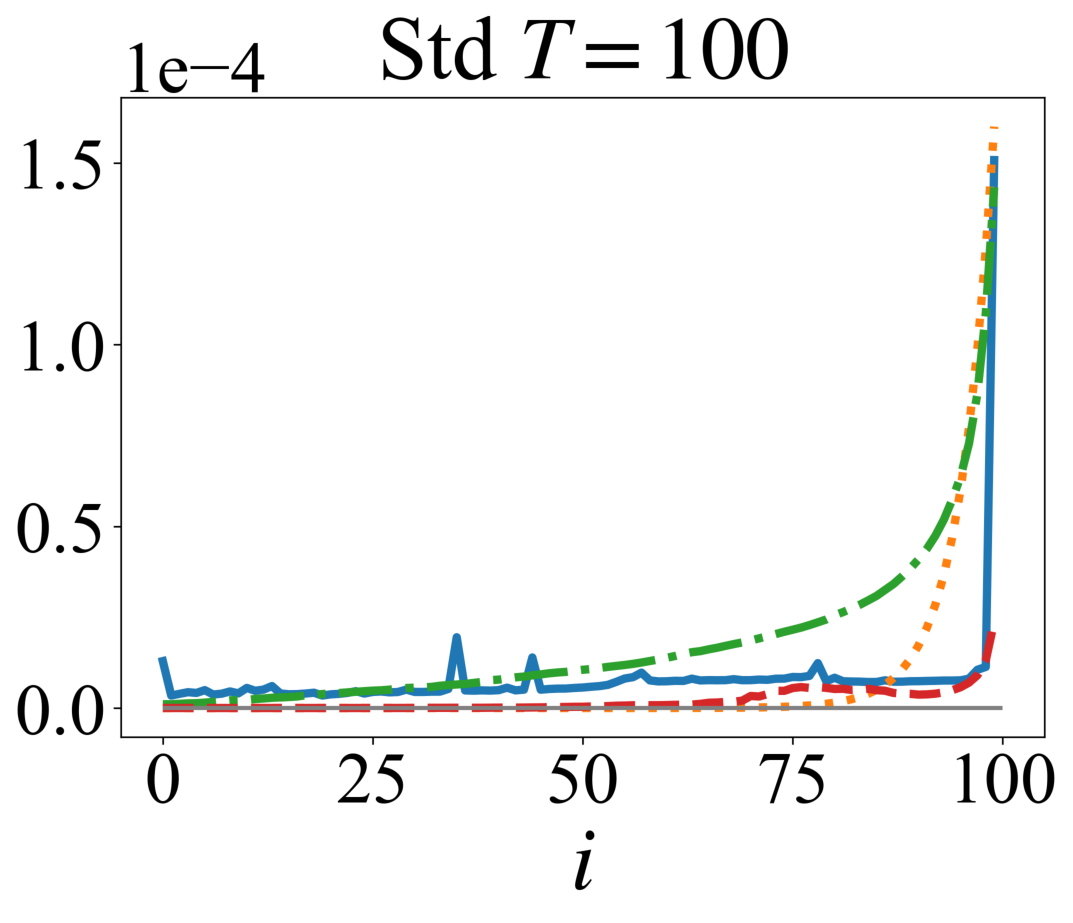}
  \end{minipage}}
  \subfigure[]{
  \begin{minipage}[b]{0.23\linewidth}
  \includegraphics[width=1\linewidth]{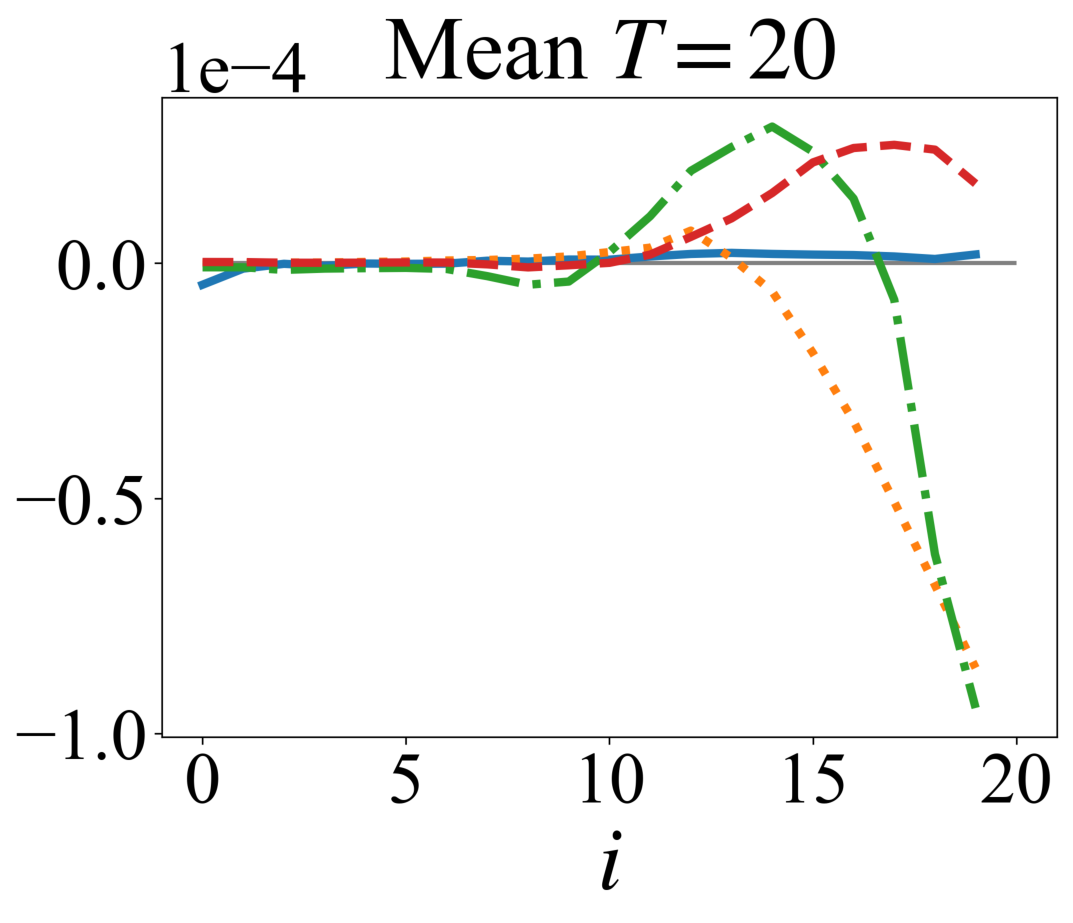}\vspace{4pt}
  \includegraphics[width=1\linewidth]{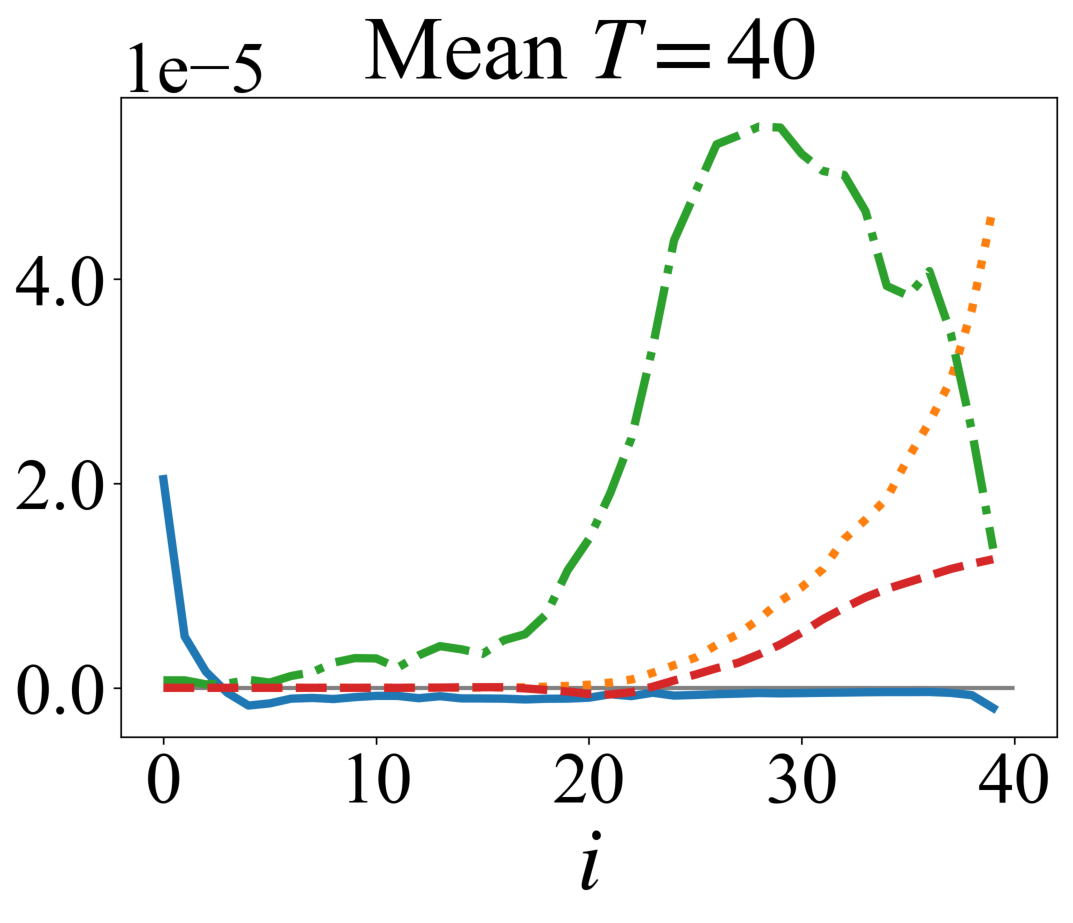}\vspace{4pt}
  \includegraphics[width=1\linewidth]{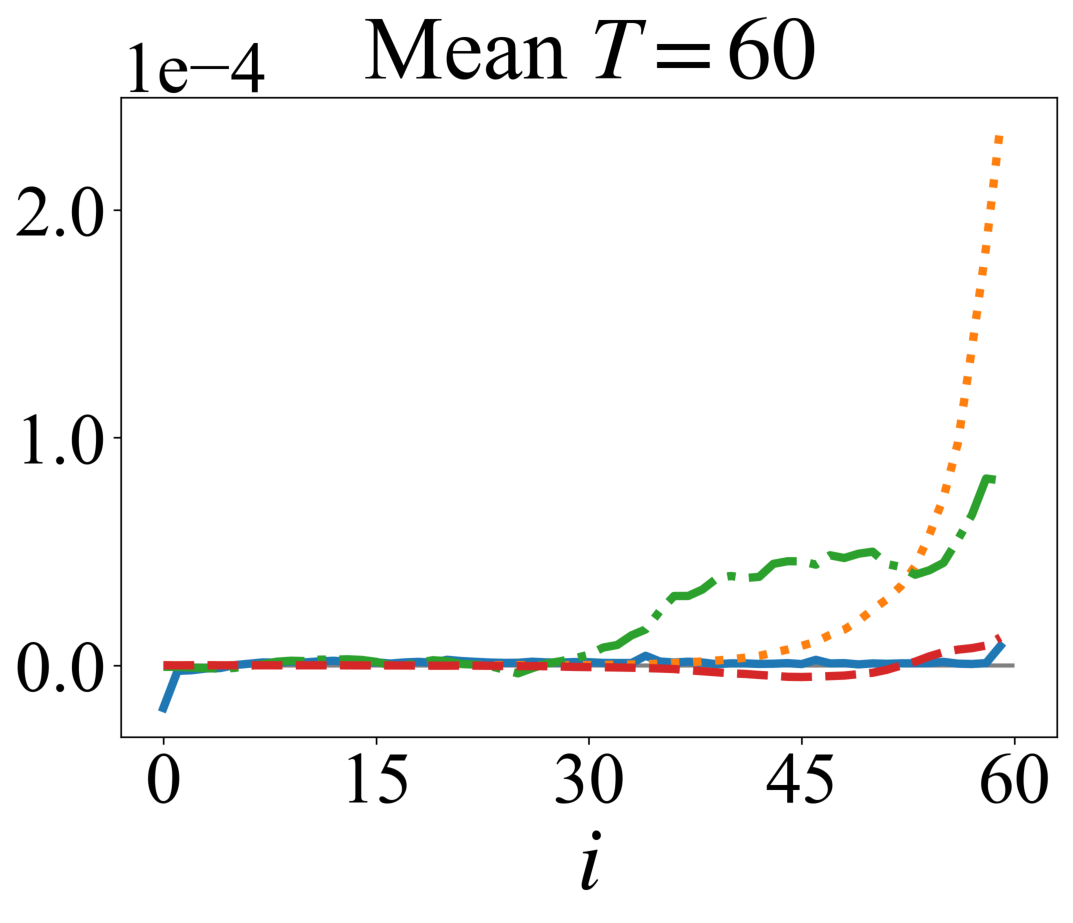}\vspace{4pt}
  \includegraphics[width=1\linewidth]{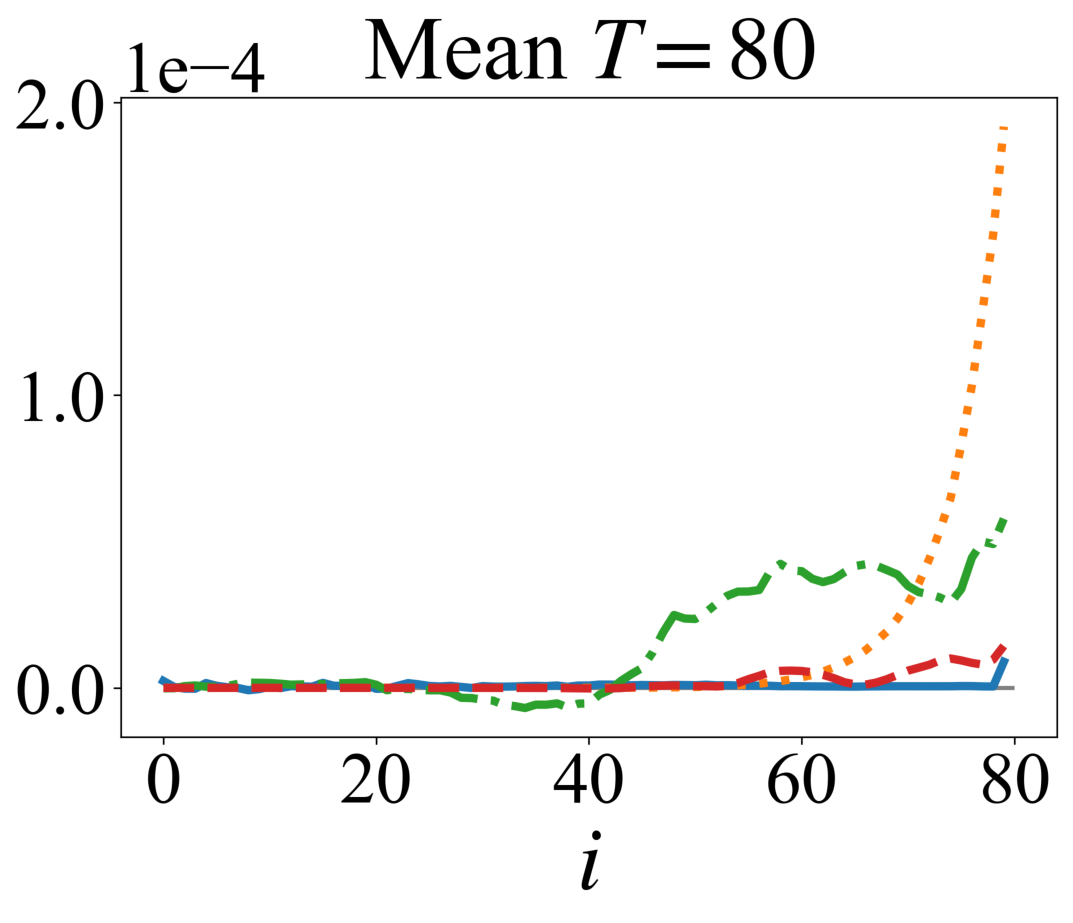}\vspace{4pt}
  \includegraphics[width=1\linewidth]{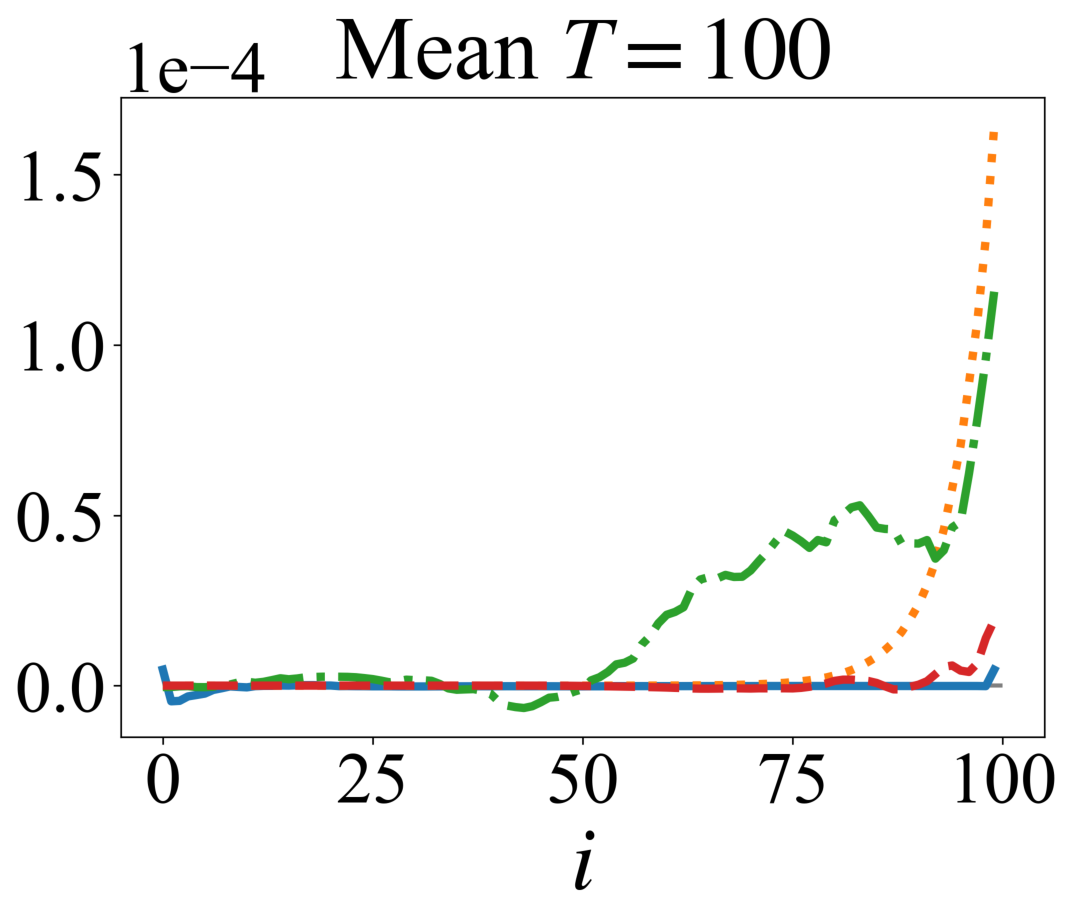}
  \end{minipage}}
  \subfigure[]{
  \begin{minipage}[b]{0.23\linewidth}
  \includegraphics[width=1\linewidth]{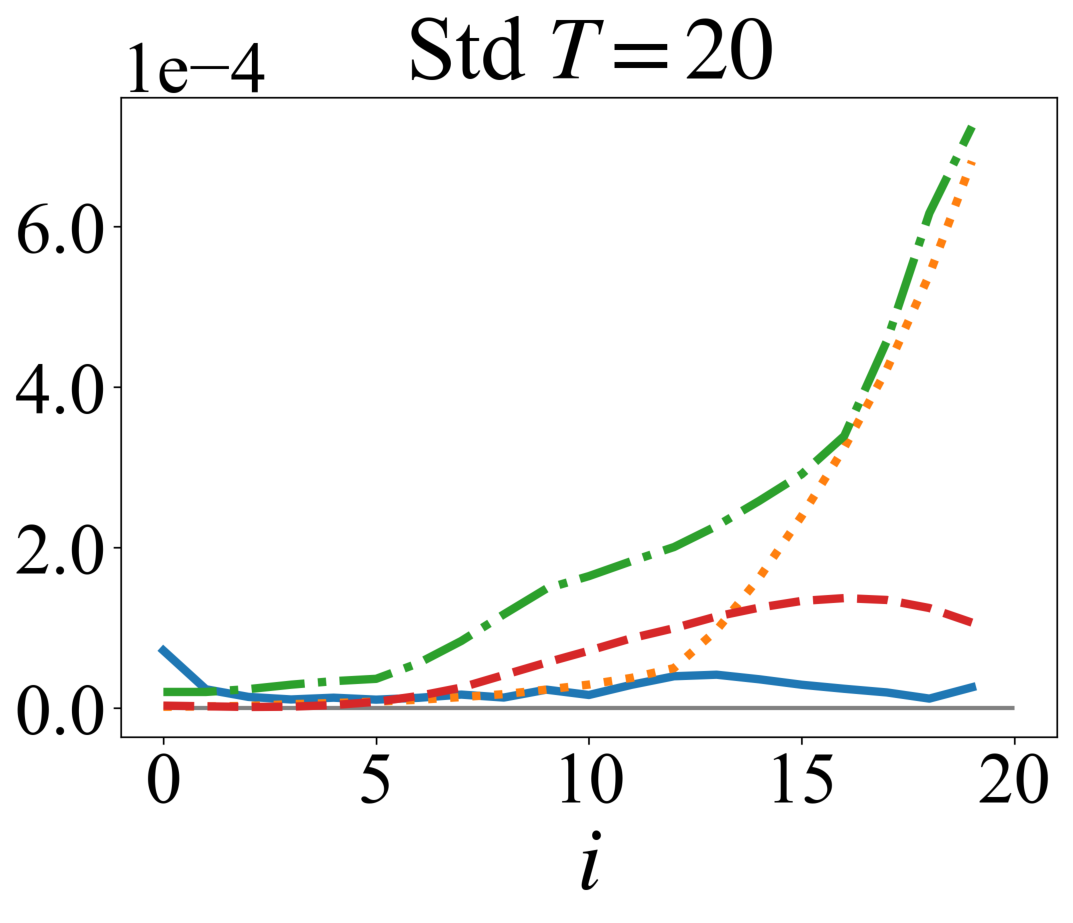}\vspace{4pt}
  \includegraphics[width=1\linewidth]{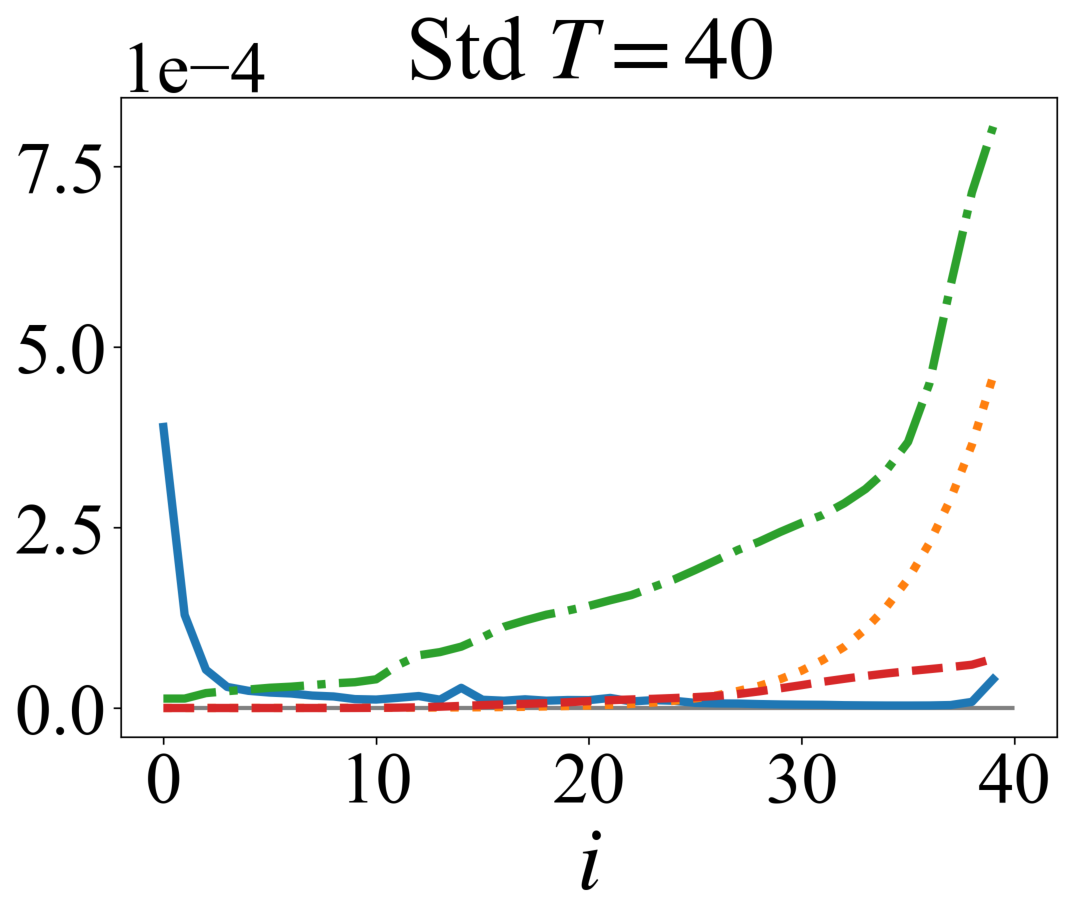}\vspace{4pt}
  \includegraphics[width=1\linewidth]{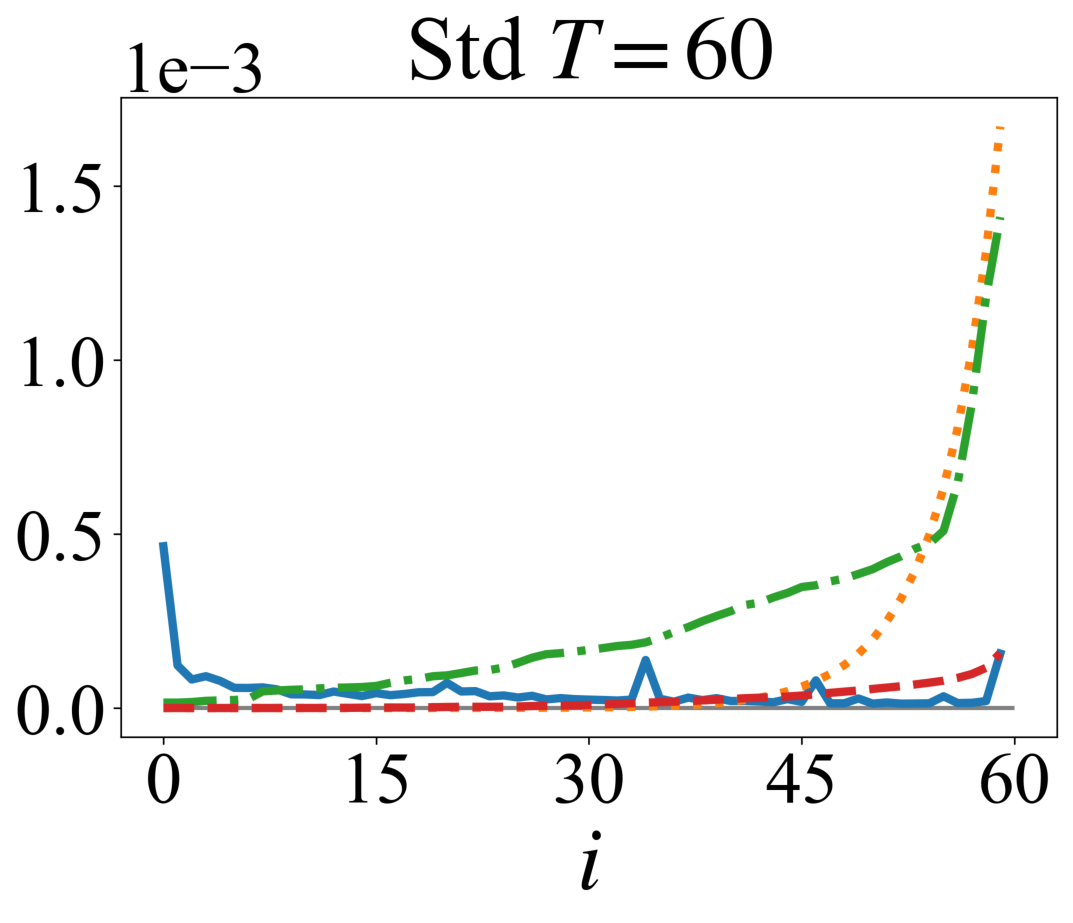}\vspace{4pt}
  \includegraphics[width=1\linewidth]{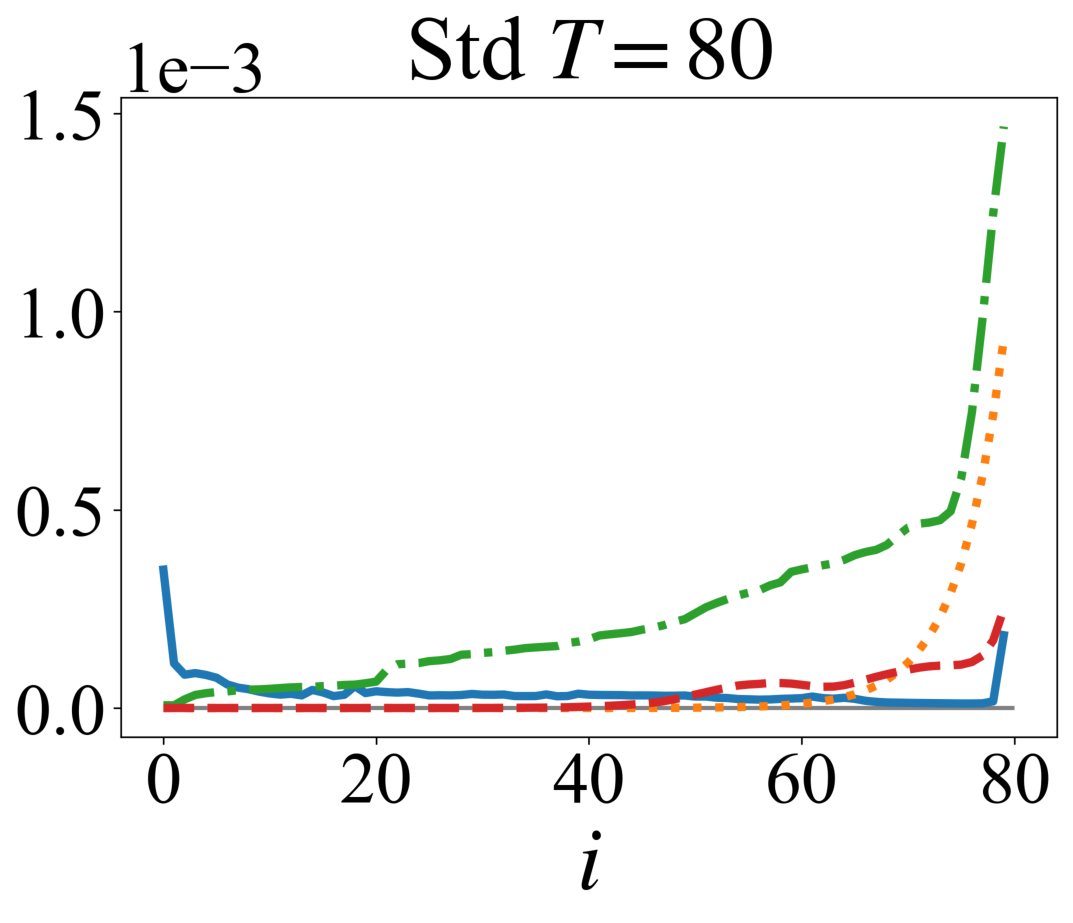}\vspace{4pt}
  \includegraphics[width=1\linewidth]{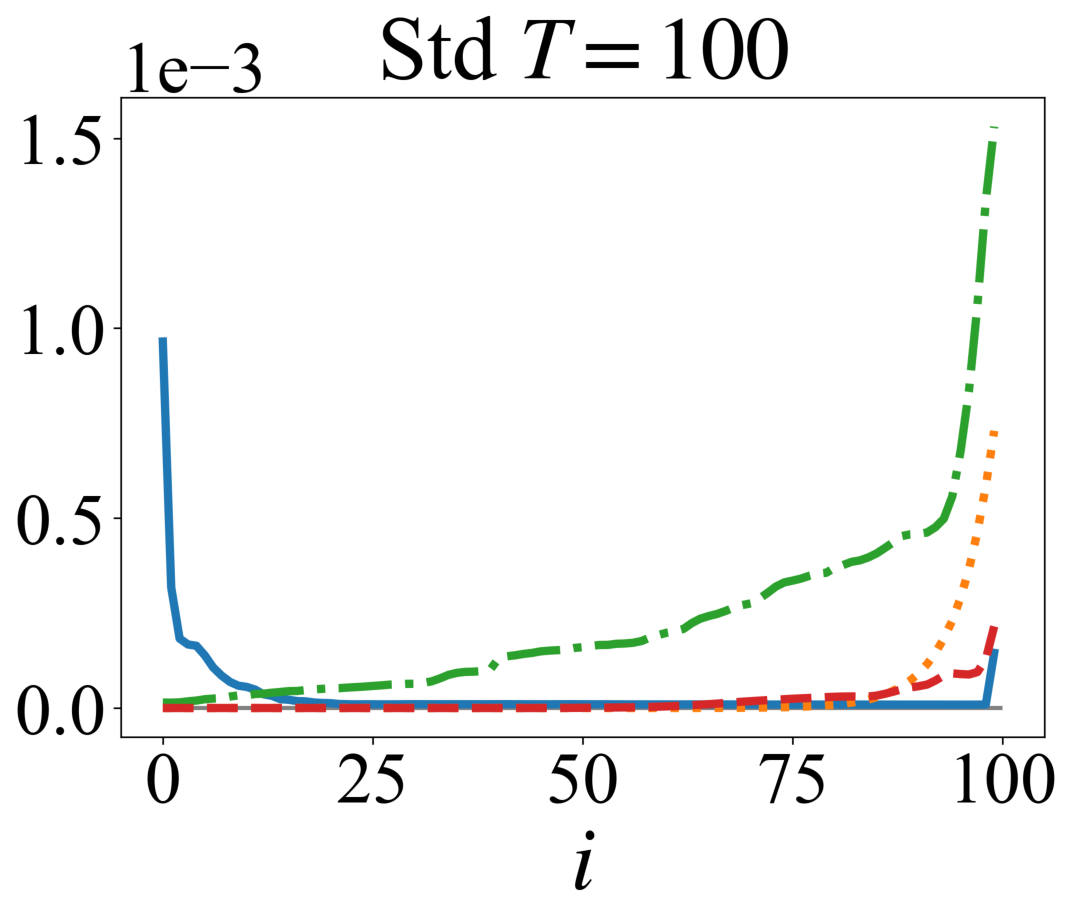}
  \end{minipage}}

  \caption{Visualization of the gradient statistics with different $T$.
  (a) The mean of gradients on SSC;
  (b) The standard deviation of gradients on SSC;
  (c) The mean of gradients on SHD;
  (d) The standard deviation of gradients on SHD.
  }
  \label{fig_grad}
\end{figure*}

We generated datasets with $T \in \{20,40,60,80,100\}$, and all networks used only one GPN layer. To avoid overfitting, we selected pre-convergence checkpoints of the training losses, and computed the gradients $\frac{\partial{\mathcal{L}}}{\partial{\bm{x}_i}} \in \mathbb{R}^{N \times C}$ of \emph{last-step-loss} $\mathcal{L}$ w.r.t to the input $\bm{x}_i$, where $N$ is the batch size and $C$ is the number of input channels. Figure~\ref{fig_grad} depicts the mean and standard deviation of $\frac{\partial{\mathcal{L}}}{\partial{\bm{x}_i}}$ at each time step $i$ ($i = 1,2,\ldots,T$).

The first row of Figures~\ref{fig_grad}(a) and (b) visualize the gradients on the SSC dataset with $T=20$. For LIF and IF, the mean and standard deviation of the gradients decayed to 0 as the time step $i$ approached 0, indicating that the gradients vanished over time. For Cuba-LIF and GPN, the values (especially the standard deviation) of the gradients did not converge to 0 when $i$ approached 0, and the same was also true for $T=40$. So there was an obvious gradient decay phenomenon over time for LIF and IF, while Cuba-LIF and GPN had a certain ability to maintain early gradients. As $T$ increased, Cuba-LIF was difficult to maintain gradients and exhibited a significant vanishing gradient phenomenon over time. However, GPN could maintain gradients at almost all time steps, even when we set $T=100$. It proves that GPN could effectively mitigate vanishing gradients while maintaining early gradients.

We also provided results on SHD in Figures~\ref{fig_grad}(c) and (d), where Cuba-LIF exhibited vanishing gradients throughout the entire process, and other experimental phenomena were similar to the above.

Note that when we set $T=100$, the retained gradients of GPN were not obvious on SSC, even approaching 0 at most time steps on SHD. It indicates that the GPN's ability to maintain gradients has an upper limit.

\begin{figure*}[htpb]      \centering
  \subfigure[]{
  \begin{minipage}[b]{0.23\linewidth}
  \includegraphics[width=1\linewidth]{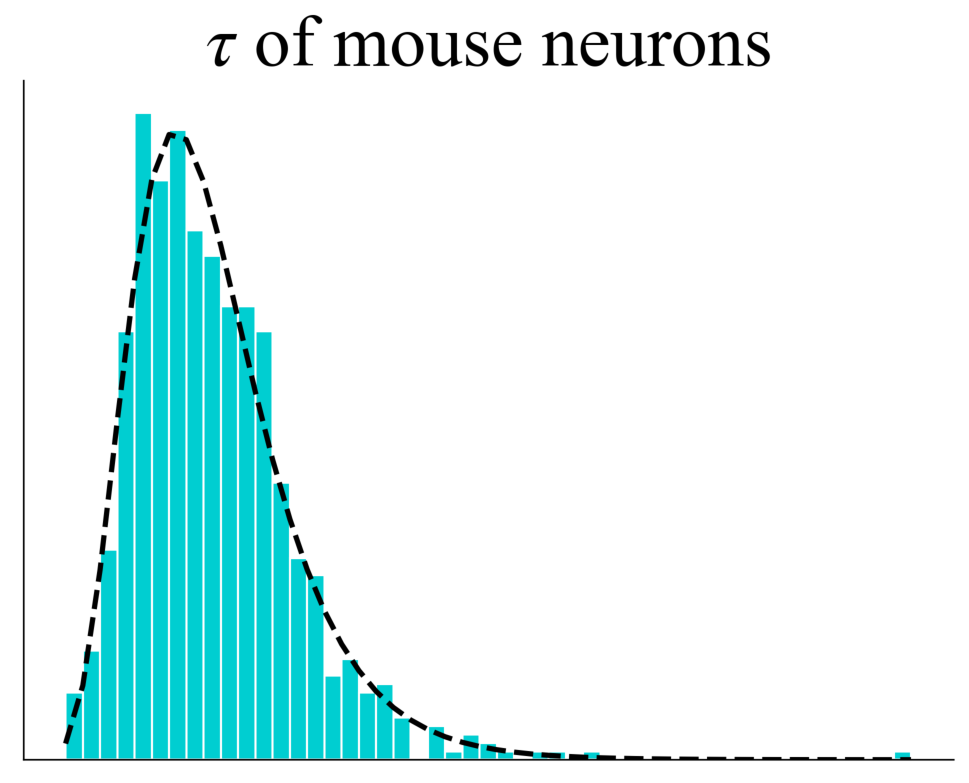}\vspace{4pt}
  \includegraphics[width=1\linewidth]{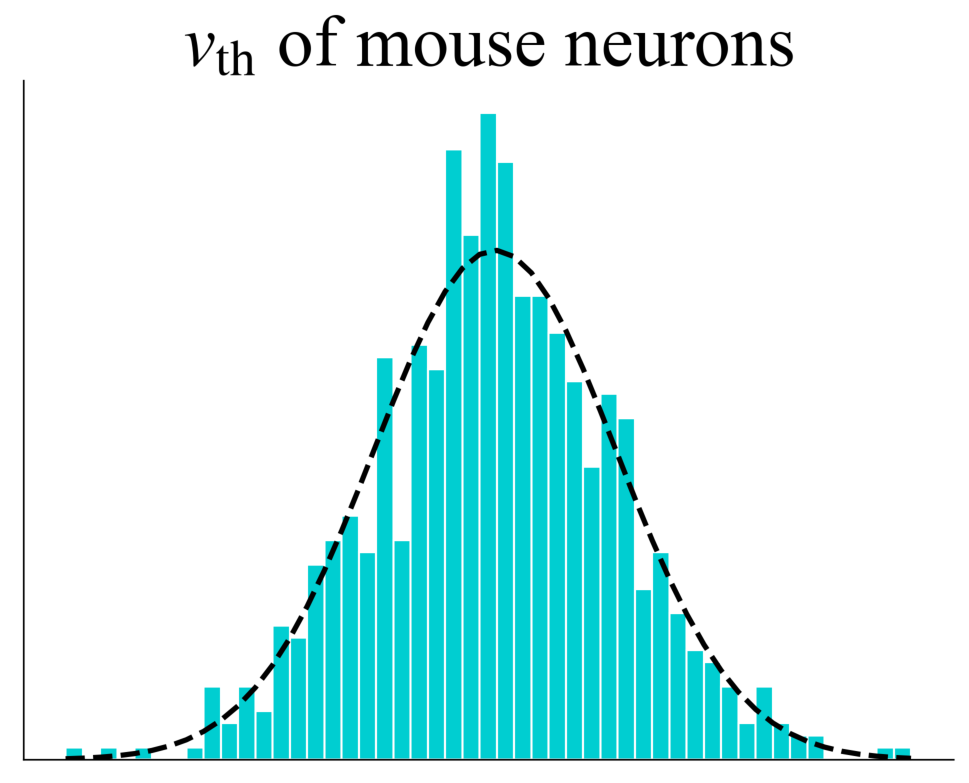}
  \end{minipage}}
  \subfigure[]{
  \begin{minipage}[b]{0.23\linewidth}
  \includegraphics[width=1\linewidth]{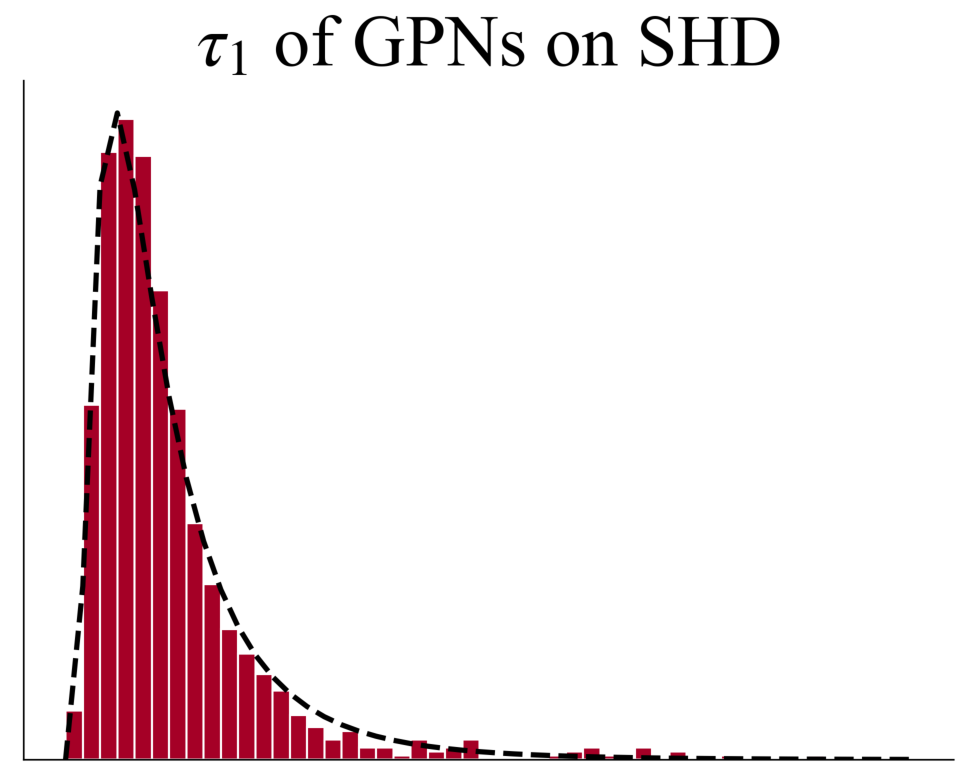}\vspace{4pt}
  \includegraphics[width=1\linewidth]{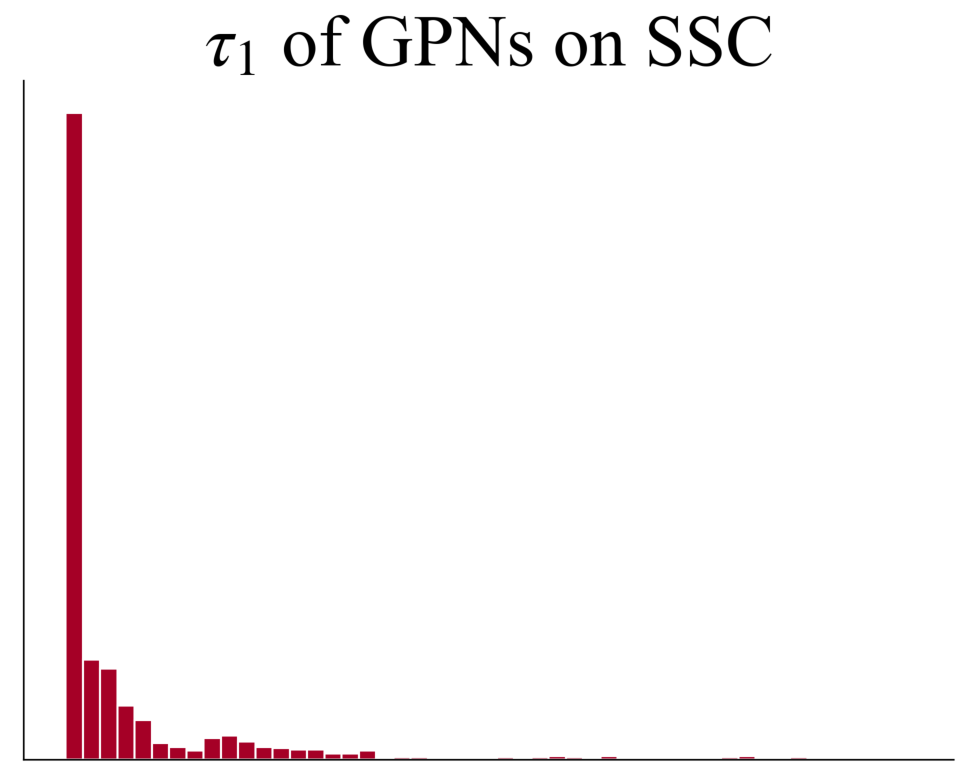}
  \end{minipage}}
  \subfigure[]{
  \begin{minipage}[b]{0.23\linewidth}
  \includegraphics[width=1\linewidth]{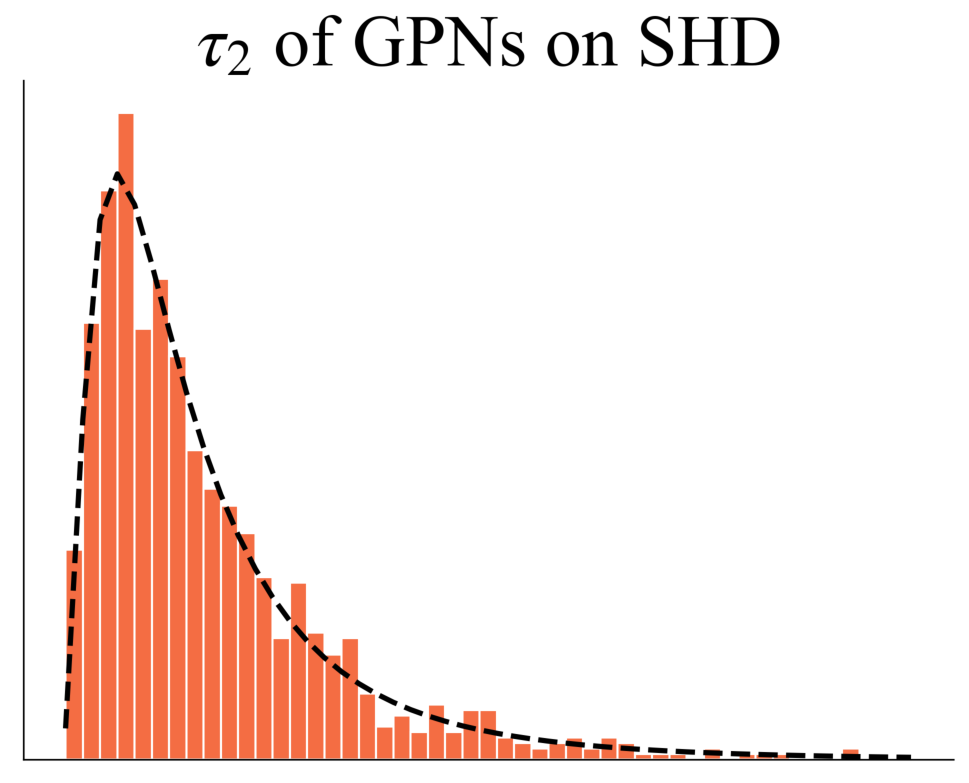}\vspace{4pt}
  \includegraphics[width=1\linewidth]{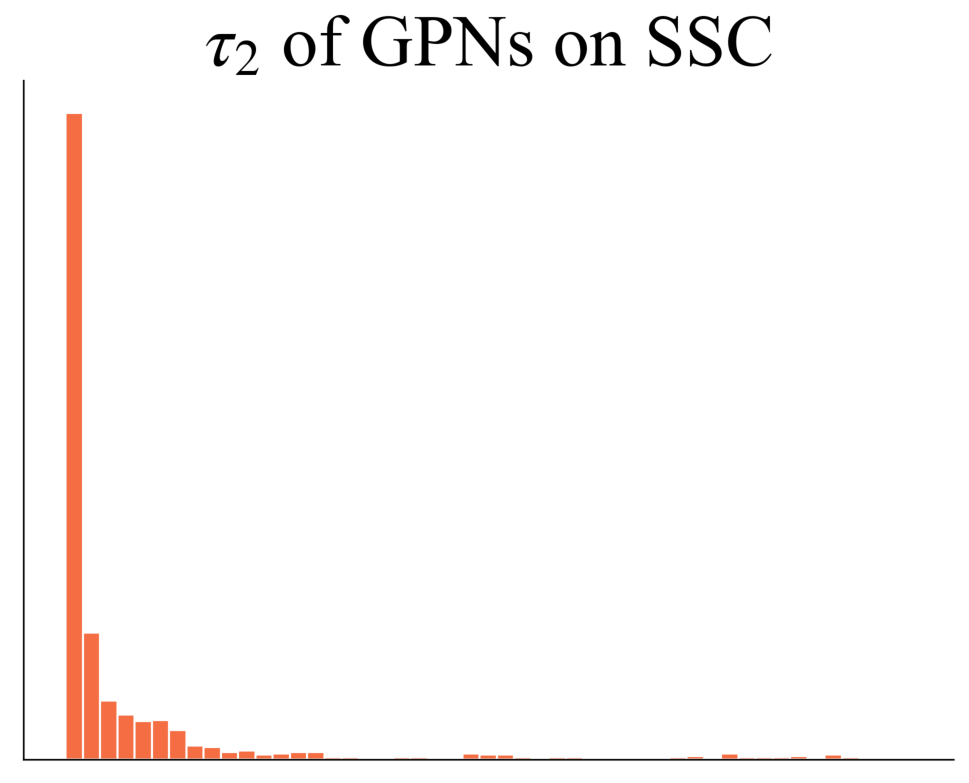}
  \end{minipage}}
  \subfigure[]{
  \begin{minipage}[b]{0.23\linewidth}
  \includegraphics[width=1\linewidth]{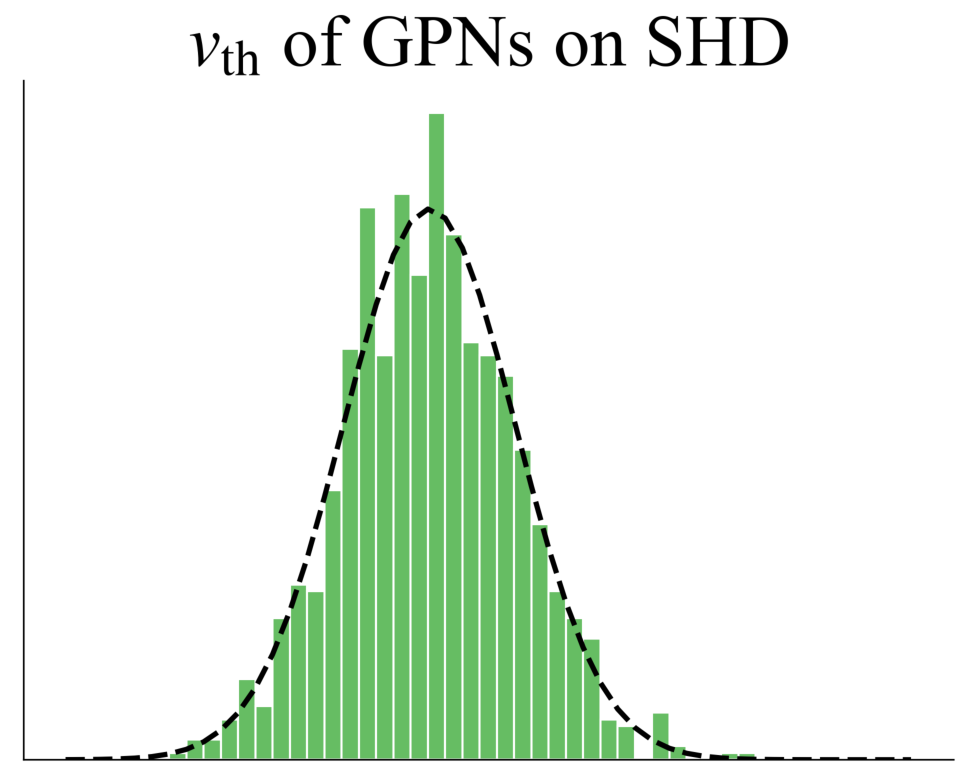}\vspace{4pt}
  \includegraphics[width=1\linewidth]{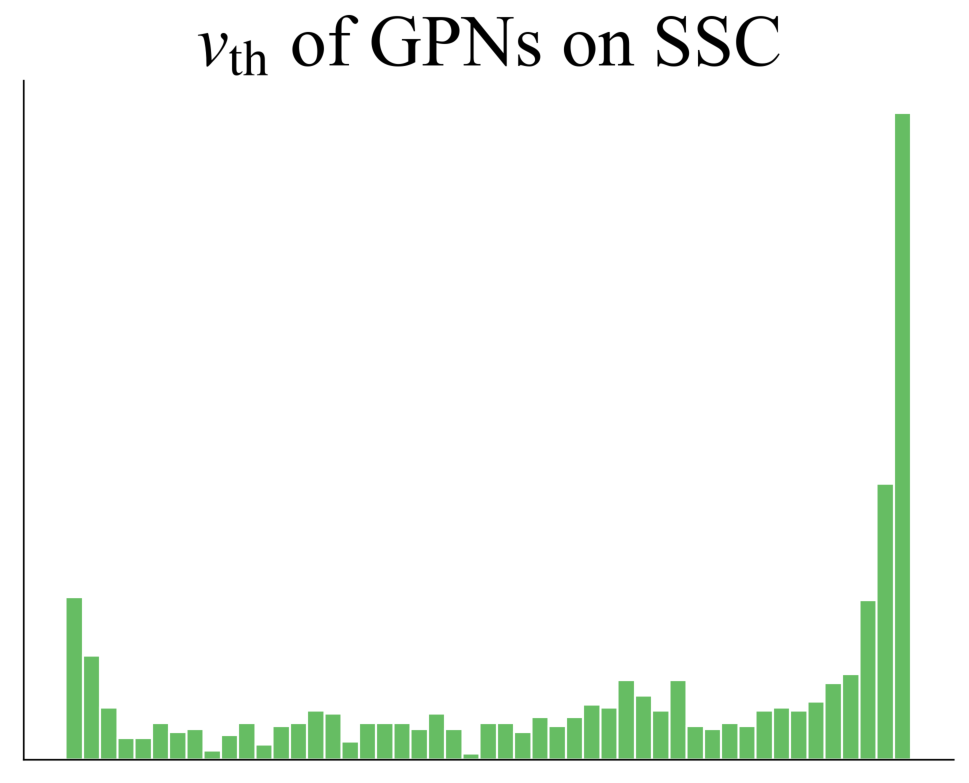}
  \end{minipage}}
  \caption{Probability density histograms of the neuronal parameters. The horizontal axis represents the parameter value, and the vertical axis represents the probability density. We focus on the shapes of distributions and omit the units. (a) Probability density histograms of the membrane time constants and firing thresholds from the mouse neurons. The firing threshold data of mouse neurons were recorded under the short square pulse stimulation. (b-c) Probability density histograms of the membrane time constants $\tau_1$ and $\tau_2$ from the GPN network on SHD and SSC. The black dotted line is the lognormal fitting curve. (d) Probability density histograms of the firing thresholds $\bm{v}_{\mathrm{th}}$ from the GPN network on SHD and SSC. The black dotted line is the normal fitting curve. }
  \label{fig_distribution}
\end{figure*}

\subsection{Spatio-temporal heterogeneous neuronal parameters}

For tasks with rich temporal structures, heterogeneous learning is more stable and robust \cite{perez2021neural}. This subsection analyzes the distributions of neuronal parameters learned by GPNs and compares them with distributions from real neurons.

We trained the GPN network on SHD and SSC training sets and extracted parameters $\widetilde{\bm{F}}$, $\widetilde{\bm{I}}$ and $\widetilde{\bm{T}}$ from a test batch. The dimensionality was $N \times C \times T$ for each type of neuronal parameter, where $N$ was the batch size, $C$ was the number of neurons in the first layer of the GPN network, and $T$ was the number of total time steps. We averaged along the dimensions of $N$ and $T$ to obtain the distributions.

Previous study \cite{perez2021neural} found that the membrane time constant distribution of mouse neurons (primary visual cortex spiny neurons in layer 4) approximates a lognormal or gamma distribution. With the same Allen Brain Atlas dataset \cite{allen2015}, Figure~\ref{fig_distribution}(a) shows that the time constant distribution approximates a lognormal distribution, whereas the threshold distribution approximates a normal distribution. For the GPN network, Figures~\ref{fig_distribution}(b) and (c) depict the histograms of the membrane time constants $\bm{\tau}_1$ and $\bm{\tau}_2$, converted respectively from leaky factors $\widetilde{\bm{F}}$ and $\widetilde{\bm{I}}$ using (\ref{eq_LIF_3}). On the SHD dataset, the distributions of $\bm{\tau}_1$ and $\bm{\tau}_2$ were similar, both approximating the lognormal distribution, whereas the distributions on SSC were more concentrated at the minimum value. In Figure~\ref{fig_distribution}(d), the threshold distribution trained on SHD approximated a normal distribution, whereas the distribution on SSC concentrated at the maximum value.

\begin{figure}[htpb]    \centering
    \subfigure{\includegraphics[width=\linewidth]{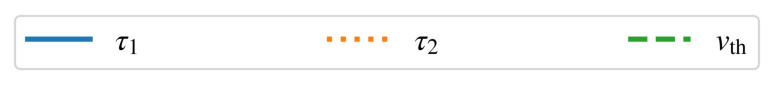}}
    \setcounter{subfigure}{0}
    \subfigure[]{\includegraphics[width=.49\linewidth,clip]{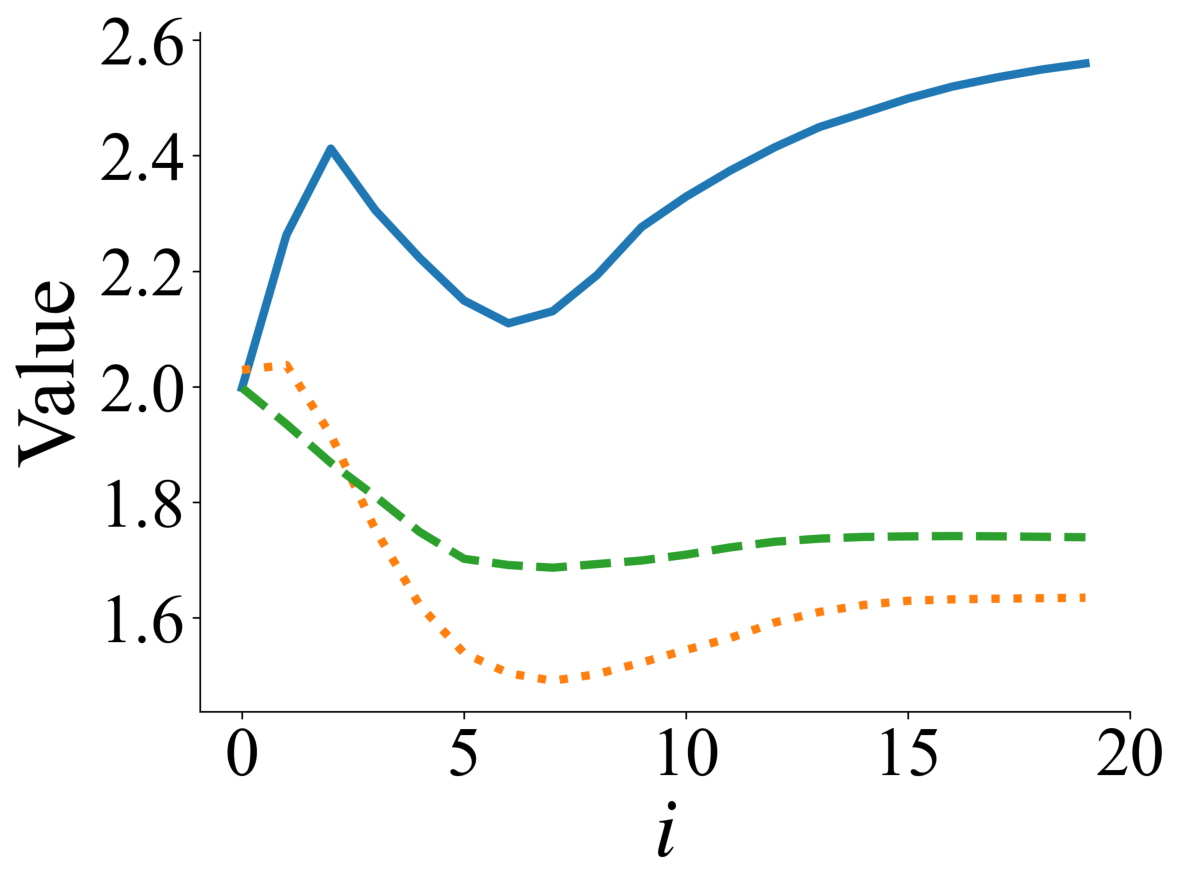}}
    \subfigure[]{\includegraphics[width=.49\linewidth,clip]{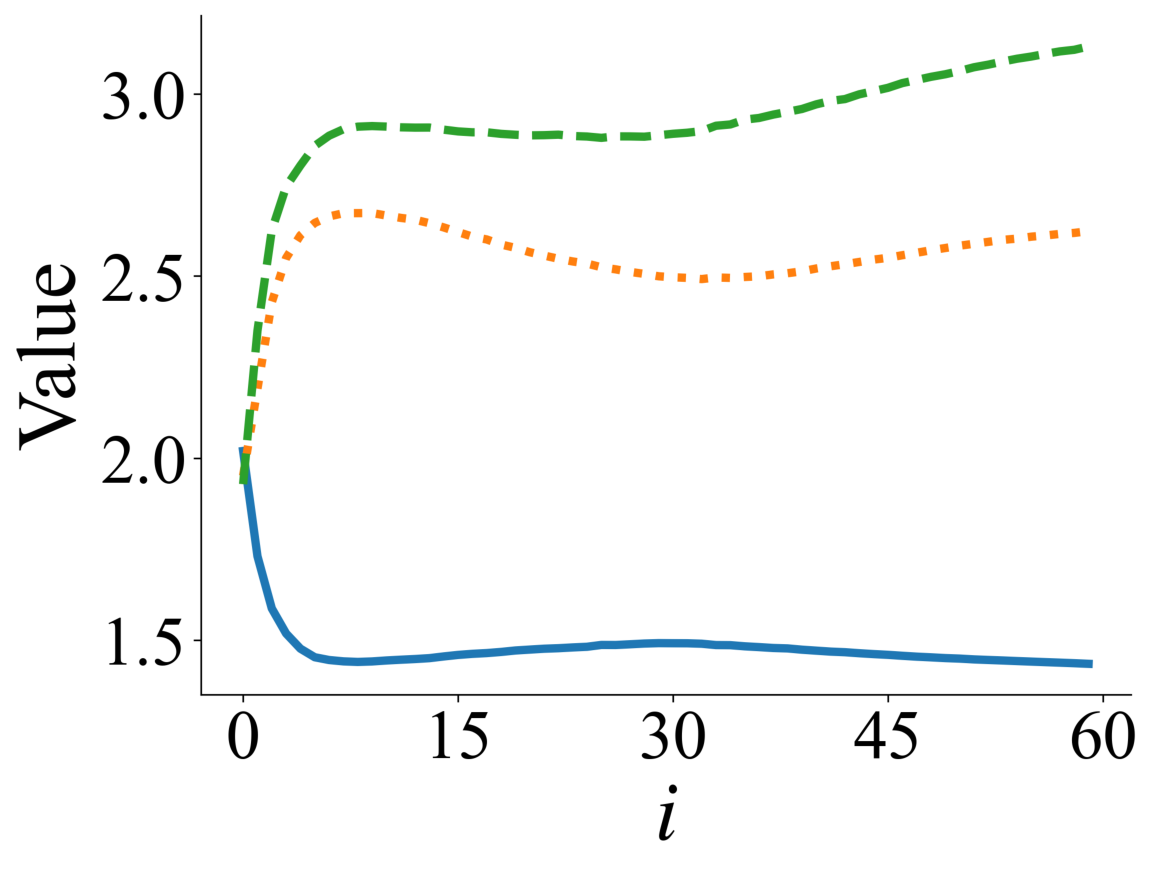}}
    \caption{Changes in neuronal parameters over time. (a) SHD; and, (b) SSC.}
    \label{fig_dynamic}
\end{figure}

In summary, the spatial distributions of neuronal parameters learned by GPN did not follow a specific fixed form. For GPN, learning neuronal parameters close to the natural distribution was not always the optimal solution. However, these results all demonstrated the spatial heterogeneity of neuronal parameters of the GPN network. Figure~\ref{fig_dynamic} shows how these neuronal parameters changed over time. Although it is difficult to analyze the pattern of changes, the temporal heterogeneity is evident.

\subsection{Experiments on other types of data}

GPN is originally designed for analyzing sequence data with high temporal complexity. This subsection studies whether it can be used on other types of data, e.g., electroencephalogram signals in brain-computer interfaces, and images from dynamic vision sensor cameras.

BCI Competition IV Dataset 2a \cite{brunner2008graz} is a commonly used motor imagery brain-computer interface dataset. We used different SNNs for motor imagery classification. The results are shown in Table~\ref{table_generalization}. Again, GPN achieved the highest classification accuracy.

DVS128Gesture \cite{Amir_2017_CVPR} is a gesture recognition dataset that used dynamic vision sensor cameras to record human gestures. Similar to ConvLSTM \cite{SHI2015Convolutional}, we modified the GPN to conduct the experiments. GPN did not perform well in image classification.

In summary, the above experiments may suggest that GPN is more suitable for sequence data, but not spatial data like images.

\begin{table}[htpb]\center
    \caption{Performance comparison of GPN with other SNNs.} \label{table_generalization}
    \renewcommand\arraystretch{1.3}
    \setlength{\tabcolsep}{12pt}{
    \begin{tabular}{cccc}        \toprule
        \textbf{Model} & \textbf{BCI2a} & \textbf{DVS128Gesture} \\
        \midrule
            IF & $37.9\pm1.1$  & $91.2 \pm 0.9$  \\
            LIF & $55.2 \pm 0.4$ & $88.0 \pm 0.5$   \\
            Cuba-LIF & $42.4 \pm 3.7$ & $91.2 \pm 0.9 $  \\
            GPN  &  $58.4 \pm 1.2$ & $84.6 \pm 2.3$  \\
     \bottomrule
    \end{tabular}}
\end{table}

\section{Conclusions} \label{sect:Conclusions}

To cope with the limitations of the LIF network, i.e., the vanishing gradient problem and fixed neuronal parameters, we have proposed GPN with gating mechanisms. To remedy the vanishing gradients of the LIF network occurring in the membrane potential pathway, we add several gates to calculate the membrane leaky factors. GPN avoids manual initialization of the neuronal parameters; moreover, its neuronal parameters demonstrate spatio-temporal heterogeneity. Experiments showed that GPN can achieve better classification performance than previous SNN approaches.

Our research explores how combining SNN with gating mechanisms can effectively resolve the limitations of SNNs, demonstrating the ability of SNN to handle long-term dependencies and achieve high classification performance simultaneously.

\section*{Acknowledgements}

This research was supported by STI 2030-Major Project 2021ZD0201300.

\section*{Declaration of Competing Interest}

The authors declare no conflict of interest.

\end{document}